\documentclass[11pt]{article}

\usepackage[final]{acl}

\usepackage[table]{xcolor}
\usepackage{times}
\usepackage{latexsym}
\usepackage{graphicx} 
\graphicspath{{Figure/}} 
\usepackage{amsfonts}
\usepackage{amsmath}   
\usepackage{amssymb}
\usepackage{enumitem} 
\usepackage{bm}       
\usepackage{subcaption}
\usepackage{booktabs}
\usepackage{multirow}
\usepackage{tabularx}
\usepackage{makecell}
\usepackage{xspace} 
\usepackage{algorithm}
\usepackage{algorithmic}
\usepackage{booktabs}
\usepackage{pifont}

\usepackage{xcolor}
\definecolor{dblue}{RGB}{0,91,150}
\definecolor{lblue}{RGB}{144,196,232}
\definecolor{gry}{RGB}{127,127,127}
\newcommand{\ourmethod}{DynHD\xspace}
\usepackage[T1]{fontenc}

\usepackage[utf8]{inputenc}

\usepackage{microtype}

\usepackage{inconsolata}

\title{\ourmethod: Hallucination Detection for Diffusion Large Language Models via Denoising Dynamics Deviation Learning}

\author{
  \textbf{Yanyu Qian\textsuperscript{1}},
  \textbf{Yue Tan\textsuperscript{2}},
  \textbf{Yixin Liu\textsuperscript{2}\textsuperscript{*}},
  \textbf{Wang Yu\textsuperscript{1}},
  \textbf{Shirui Pan\textsuperscript{2}}
\\
\\
  \textsuperscript{1}Nanyang Technological University, Singapore
  \textsuperscript{2}Griffith University, Australia 
\\
  \small{
    \textsuperscript{*}Corresponding author:
    \href{mailto:yixin.liu@griffith.edu.au}{yixin.liu@griffith.edu.au}
  }
}

\begin{document}
\maketitle

\begin{abstract}
Diffusion large language models (D-LLMs) have emerged as a promising alternative to auto-regressive models due to their iterative refinement capabilities. 
However, hallucinations remain a critical issue that hinders their reliability. 
To detect hallucination responses from model outputs, token-level uncertainty (e.g., entropy) has been widely used to indicate potential factual errors. 
Nevertheless, unlike auto-regressive models that generate tokens sequentially, D-LLMs generate fixed-length sequences simultaneously, where only a small subset of tokens is informative for hallucination detection. Thus, aggregating uncertainty over all tokens can be suboptimal. 
Moreover, the evolution trend of uncertainty throughout the diffusion process can also provide valuable signals, highlighting the necessity of modeling its denoising dynamics for hallucination detection. 
In this paper, we propose \ourmethod that bridges these gaps from both \textit{spatial} (token sequence) and \textit{temporal} (denoising dynamics) perspectives. 
To handle the information density imbalance across tokens, we propose a semantic-aware evidence construction module that extracts hallucination-indicative signals by removing task-invariant structural tokens and emphasizing the uncertainty of the remaining informative tokens. 
To model denoising dynamics for hallucination detection, we introduce a reference evidence generator that learns the expected evolution trajectory of uncertainty evidence, along with a deviation-based hallucination detector that makes predictions by measuring the discrepancy between the observed and reference trajectories. 
Extensive experiments demonstrate that \ourmethod consistently outperforms state-of-the-art baselines while achieving higher efficiency across multiple benchmarks and backbone models. The code is available at: \url{https://github.com/qyy11-com/DynHD}.
\end{abstract}

\section{Introduction}\label{sec:introduction}

Diffusion large language models (D-LLMs)~\citep{nie-etal-2025-llada,ye-etal-2025-dream} have recently received extensive attention as a significant generative paradigm alternative to autoregressive large language models (AR-LLMs). 
Unlike AR-LLMs using token-by-token decoding, D-LLMs generate fixed-length sequences through multi-step iterative denoising, performing discrete remasking at each step: the model predicts a denoised intermediate sequence while selectively remasking a portion of positions for further optimization~\cite{austin-etal-2021-structured,zhao2026fedcigar}. 
This paradigm offers advantages in global consistency and the ability to gradually refine the generated content~\cite{li-etal-2022-diffusionlm,huang-etal-2024-trustllm}.

\begin{figure}[t] 
    \centering
    \includegraphics[width=\columnwidth]{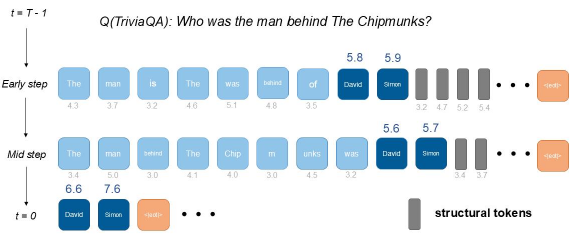}
    \caption{Visualization of spatial uncertainty distribution during decoding. Tokens exhibiting the \textcolor{dblue}{highest entropy spikes} serve as the primary indicators of factual instability. On the contrary, \textcolor{gry}{structural} tokens provide limited cues for hallucination detection. }
    \label{fig:denoising_token}
\end{figure}

\begin{figure*}[t]
    \centering
    \begin{subfigure}[b]{0.32\textwidth}
        \centering
        \includegraphics[width=\textwidth]{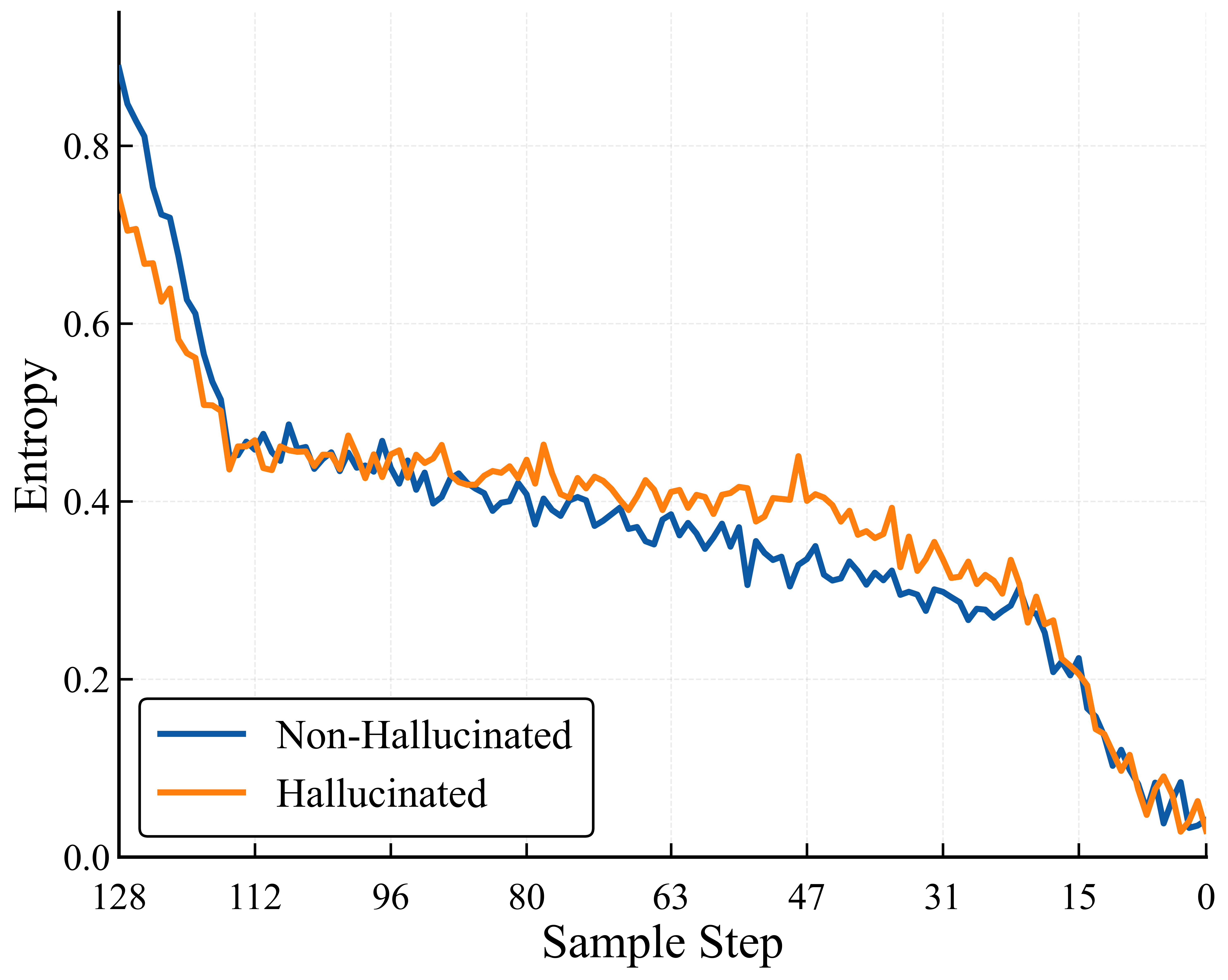}
        \caption{}
        \label{fig:mean_entropy}
    \end{subfigure}
    \hfill
    \begin{subfigure}[b]{0.32\textwidth}
        \centering
        \includegraphics[width=\textwidth]{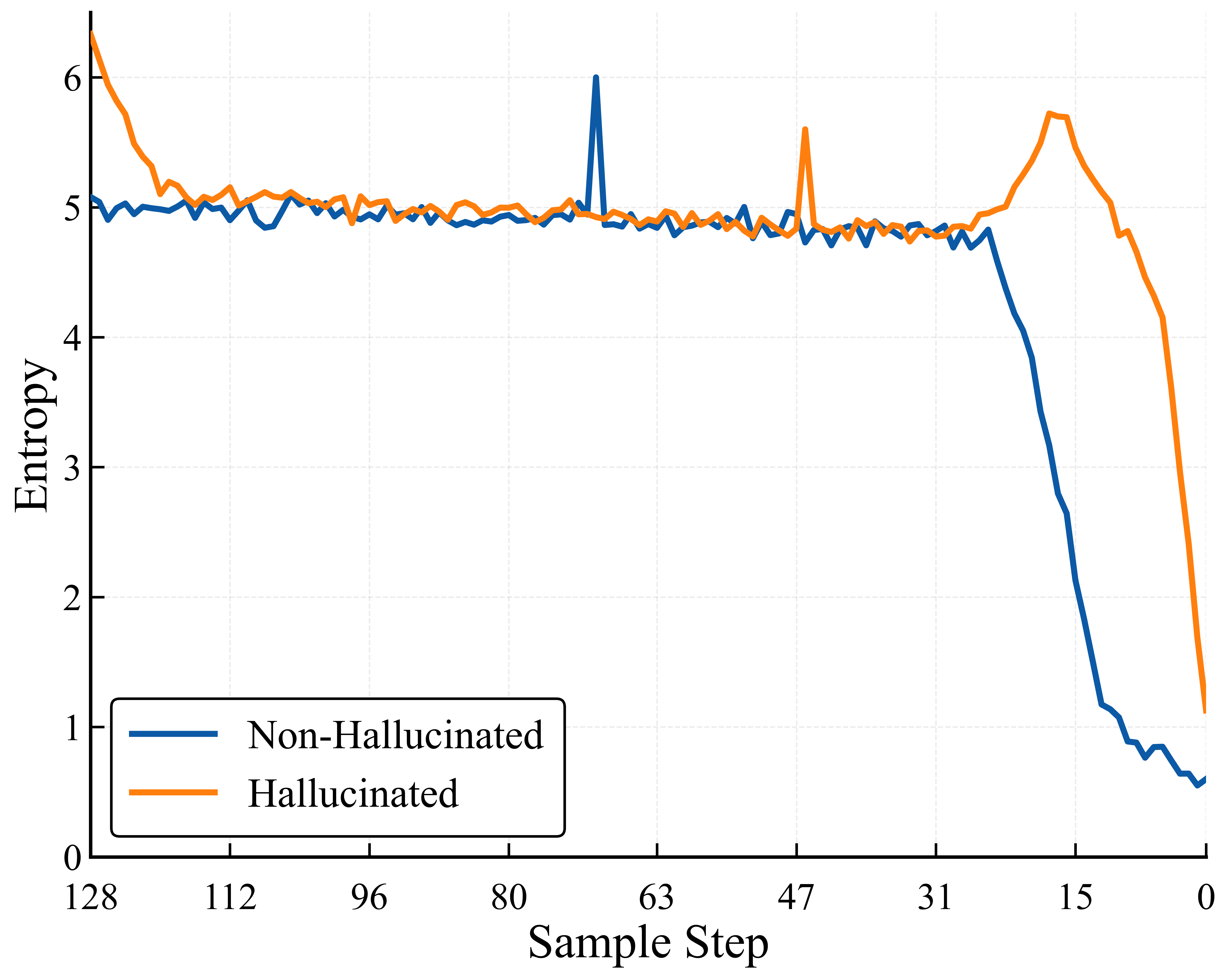}
        \caption{}
        \label{fig:hfig1}
    \end{subfigure}
    \hfill
    \begin{subfigure}[b]{0.32\textwidth}
        \centering
        \includegraphics[width=\textwidth]{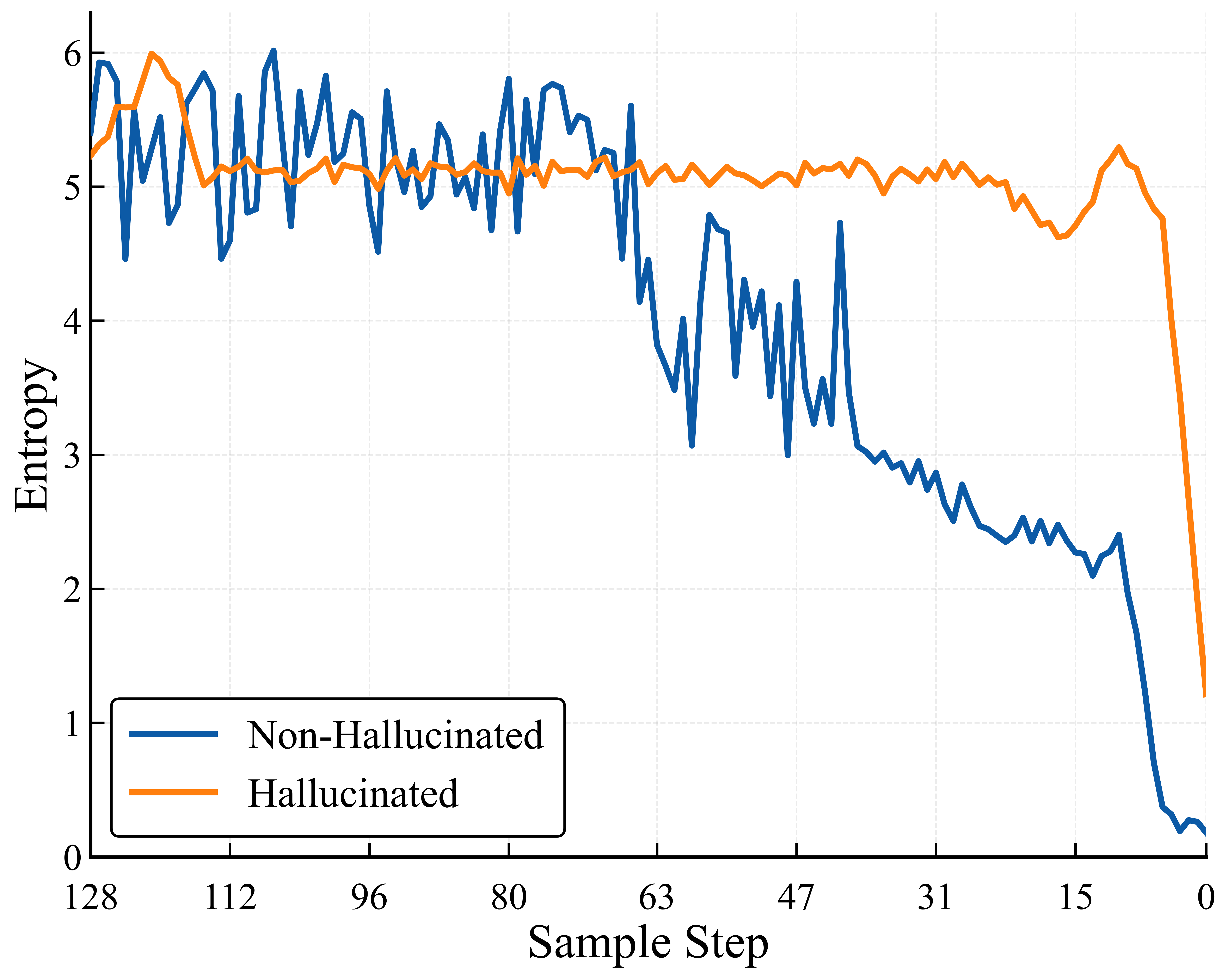}
        \caption{}
        \label{fig:tfig1}
    \end{subfigure}
    \caption{Entropy dynamics under spatial filtering. (a) Mean trajectories fail to distinguish between samples due to {information density imbalance}. (b) \& (c) Filtered trajectories (top-$K$) reveal distinct convergence: factual samples decay monotonically, while hallucinated ones exhibit late-stage stagnation or rebound. Differences in decay rates between HotpotQA (b) and TriviaQA (c) reflect varying task-specific denoising dynamics.}
    \label{fig:combined_dynamics}
\end{figure*}
While D-LLMs offer several advantages, they still suffer from hallucinations, i.e., generating factually incorrect or unsupported content, which limits their reliability in real-world applications~\cite{ji2025denoising,liu2026beyond}. To mitigate this issue, hallucination detection has become a critical task for ensuring the factual reliability of D-LLMs. Nevertheless, although extensive efforts have been devoted to the AR-LLMs setting, these hallucination detection approaches do not readily generalize to D-LLMs due to their fundamentally different architectural and generative paradigms~\cite{yehuda2024interrogatellm,tan2026influence,li2026towards}. In particular, the key evidence identifying hallucinations in D-LLMs is typically distributed across the dynamic evolution of the entire denoising trajectory, rather than being concentrated in the output confidence of the final step. 
Consequently, hallucination detectors designed for AR-LLMs, such as those relying on final token probabilities~\cite{kossen-etal-2024-semantic}, single-step uncertainty and entropy~\cite{du-etal-2024-haloscope}, or multi-sample consistency~\cite{manakul-etal-2023-selfcheckgpt}, often ignore the process-level signals inherent in the denoising trajectory. Moreover, these approaches, which are already computationally expensive for AR-LLMs, become even more time-consuming for D-LLMs, as additional sampling or verification requires repeatedly executing the denoising process~\cite{chern-etal-2023-factool}.

As a pioneering work on hallucination detection for D-LLMs, TraceDet~\cite{chang-etal-2025-tracedet} adopts a trajectory-based perspective by treating the denoising process as observable action traces.
It extracts hallucination-relevant sub-trajectories for discrimination and emphasizes the importance of ``process signals'' over ``final-state signals'' for hallucination detection in D-LLMs. 
However, through our in-depth analysis of the hallucination-indicated signals in D-LLMs, TraceDet still exhibits two orthogonal weaknesses: 
From a spatial perspective, it ignores the  \textbf{information density imbalance} within generated sequences, failing to explicitly filter out structural padding tokens introduced by fixed-length generation. From a temporal perspective, it overlooks the underlying \textbf{denoising dynamics} of entropy-based hallucination evidence, hindering the modeling of uncertainty evolution in the diffusion trajectory. 

Firstly, regarding the spatial dimension, D-LLMs naturally generate answers within fixed-length sequences~\cite{li-etal-2022-diffusionlm,chen2025multi}. In these sequences, the roles of individual tokens are highly heterogeneous, leading to an \textbf{information density imbalance}~\cite{gong-etal-2022-diffuseq,tan2023taming}, as illustrated in Figure~\ref{fig:denoising_token}. A few tokens are vital for the final answer, providing key signals for hallucination detection. In contrast, many positions are filled with structural padding tokens introduced by fixed-length generation, which carry little hallucination-related information and should be explicitly filtered out.

In the context of such imbalance, TraceDet~\cite{chang-etal-2025-tracedet} aggregates uncertainty from all tokens as detection evidence without explicitly accounting for their semantic importance, which may drown out hallucination signals with low-informative tokens~\cite{zhang-etal-2023-snowball,pan2026camera}. As a result, the entropy profiles of hallucinated and factual samples become indistinguishable from each other (see Figure~\ref{fig:mean_entropy}), leading to sub-optimal hallucination detection performance.

Even if hallucination-related evidence can be extracted from token sequences, how to leverage the evidence trajectory over multiple diffusion steps remains another challenge. Through visualization of evidence trajectories, we empirically find that \textbf{denoising dynamics}, i.e., the trend of how evidence evolves throughout the diffusion process, are critical for hallucination detection. As shown in Figure~\ref{fig:hfig1} and Figure~\ref{fig:tfig1} (corresponding examples are in Appendix~\ref{sec:appendix_case_studies}), factual and hallucinated samples often show different trends during the denoising process, providing strong discriminative signals. More importantly, the range, trend, and shape of entropy curves often vary significantly across different scenarios and questions (see the difference between Figures~\ref{fig:hfig1} and~\ref{fig:tfig1}), highlighting the necessity of modeling complex denoising dynamics for robust hallucination detection. 
Nevertheless, TraceDet selects only a few key steps for hallucination prediction, which may neglect the continuous and holistic dynamics of the diffusion process. 

To overcome the above limitations, we propose \textbf{\ourmethod}, a {denoising \textbf{Dyn}amics deviation learning}-based approach for \textbf{H}allucination \textbf{D}etection of D-LLMs. 
To address the \textit{information density imbalance}, our method introduces a semantic-aware token filtering module to explicitly remove structural padding tokens introduced by fixed-length generation, and further leverages multivariate statistical modeling of entropy to construct evidence with strong hallucination indications.
Moreover, we model denoising dynamics for robust hallucination detection with a denoising dynamical deviation learning module, where a reference evidence dynamics generator is established to model the normal evolution patterns of evidence, and a deviation-based detector identifies hallucinations by measuring the difference between observed evidence dynamics and the reference dynamics. 
 
Extensive experiments on TriviaQA~\cite{joshi-etal-2017-triviaqa}, HotpotQA~\cite{yang-etal-2018-hotpotqa}, and CommonsenseQA~\cite{talmor-etal-2019-commonsenseqa} with Dream-7B~\cite{ye-etal-2025-dream} and LLaDA-8B~\cite{nie-etal-2025-llada} demonstrate that \ourmethod achieves state-of-the-art performance. It outperforms TraceDet by an average AUROC margin of 12.2\% without requiring external retrieval or computationally expensive repeated sampling.

\section{Preliminaries}\label{sec:preliminaries}

\paragraph{Diffusion Denoising Mechanism.}
Diffusion large language models (D-LLMs) generate sequences via an iterative stochastic denoising process. Given a prompt $\mathbf{q}$, the model maintains a discrete state $\mathbf{x}^{(t)} = (x^{(1)}_t, \dots, x^{(l)}_t) \in \mathcal{V}^l$ across the trajectory $t \in \{T, \dots, 0\}$, where $\mathcal{V}, l, T$ denote the vocabulary, fixed sequence length, and total denoising steps, respectively. Starting from a highly masked initialization $\mathbf{x}^{(T)}$, the model iteratively fills masked positions to reach the final output $\mathbf{x}^{(0)}$.

At each step $t$, the model predicts a categorical distribution ${\pi}_i^{(t)} \in \mathbb{R}^{|\mathcal{V}|}$ for each position $i$ via a softmax over the generated logits. We quantify the position-wise prediction uncertainty using the token entropy:
\begin{equation}
    s_{t,i} = -\sum_{v \in \mathcal{V}} \pi_i^{(t)}(v) \log \pi_i^{(t)}(v).
\end{equation}
Let $\mathcal{S}_t = \{s_{t,i}\}_{i=1}^l$ represent the sequence of token entropies at step $t$. The complete raw uncertainty trajectory spanning all descending denoising steps is thus denoted as $\mathcal{T} = (\mathcal{S}_T, \mathcal{S}_{T-1}, \dots, \mathcal{S}_0)$. Notably, due to the fixed length $l$, sequences that terminate early are padded with non-semantic structural tokens ($\langle|endoftext|\rangle$), which introduces significant noise into $\mathcal{T}$.

\paragraph{Definition of Hallucination Detection Problem.} 
We formulate hallucination detection as a binary classification task for a question-response pair $(\mathbf{q}, \mathbf{r})$, where $\mathbf{r} = \mathbf{x}^{(0)}$. Given a training set $\mathcal{D} = \{(\mathbf{q}_n, \mathbf{r}_n, y_n)\}_{n=1}^N$ with labels, where $y=1$ indicates a hallucinated response and $y=0$ denotes a factually correct one. Our goal is to learn an optimal detector $h_{\phi}$ (e.g., a neural network parameterized by $\phi$) that minimizes the empirical risk:
\begin{equation}
\begin{aligned}
    \min_{\phi} & \quad \mathbb{E}_{(\mathbf{q},\mathbf{r},y) \sim \mathcal{D}} \left[ \mathcal{L}(y, h_{\phi}(\mathbf{q}, \mathbf{r}, \mathcal{T})) \right], \\
    \text{s.t.} & \quad \mathbf{x}^{(t-1)} \sim p_{\theta}(\mathbf{x}^{(t-1)}|\mathbf{x}^{(t)}, \mathbf{q}),
\end{aligned}
\end{equation}
where $\mathcal{L}$ is a classification loss function, and the detector makes its decision based on the question, the final response, and the collected uncertainty trajectory $\mathcal{T}$.

The related work is summarized in Appendix~\ref{sec:appendix_rw}.
\begin{figure*}[t]
    \centering
    \includegraphics[width=\textwidth]{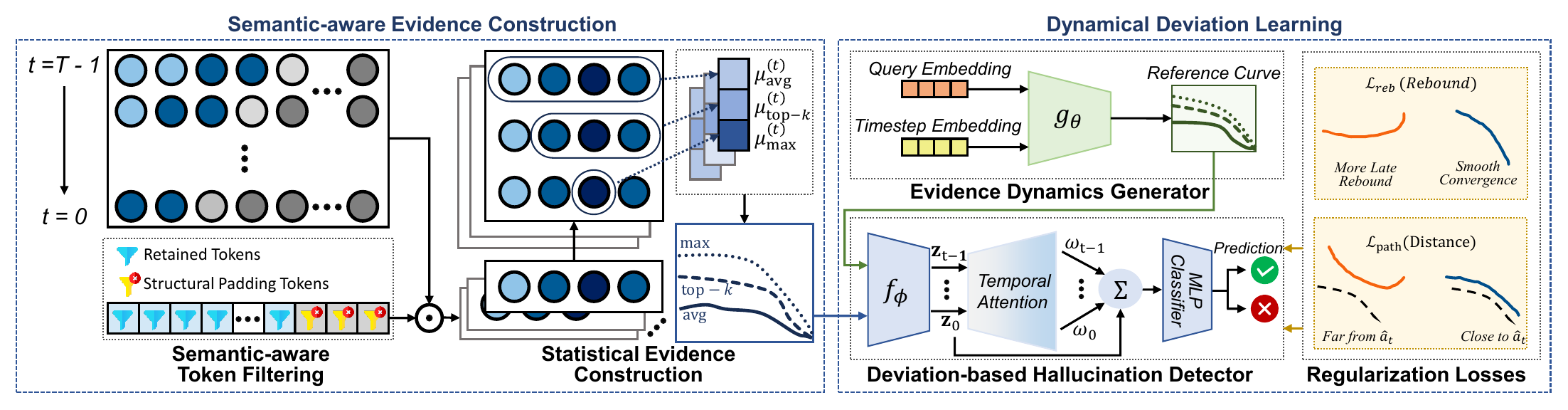}
    \caption{Framework overview of \ourmethod, which constructs a semantic-aware evidence trajectory from uncertainty, and detects hallucination via learning the deviation between the reference and observed trajectories.}
    \label{fig:framework}
\end{figure*}

\section{Methodology}\label{sec:methodology}

In this section, we introduce \ourmethod that detects hallucinations by learning deviations in denoising dynamics. 
As illustrated in Figure~\ref{fig:framework}, \ourmethod starts with a \textit{semantic-aware evidence construction} module~(Sec.~\ref{sec:evidence}). To handle the {information density imbalance} issue, it first filters out structural padding tokens introduced by fixed-length generation, and then constructs the hallucination evidence by statistically aggregating the uncertainty of the remaining tokens.
After that, a \textit{dynamical deviation learning} module~(Sec.~\ref{sec:traj_learning}) identifies the hallucination answers by modeling denoising dynamics of D-LLMs. Concretely, 
we model the continuous denoising process by building 
a question-conditioned reference path to serve as a baseline, and then compare the actual evidence trajectory against this established reference for hallucination prediction.

\subsection{Semantic-aware Evidence Construction}\label{sec:evidence}

To identify hallucinations generated by D-LLMs, the first step is to construct hallucination-indicative evidence from the denoising process. While token-level uncertainty (e.g., entropy) can serve as a useful indicator, the fixed-length generation paradigm of D-LLMs introduces an information density imbalance, i.e., not all tokens in a generated sequence provide meaningful signals for hallucination detection. To address this issue, we adopt a two-stage strategy for evidence construction. First, we conservatively filter out task-invariant structural padding tokens introduced by fixed-length generation. Then, we construct statistical evidence by aggregating the entropy of the remaining tokens.

\paragraph{Semantic-aware Token Filtering.}

Due to the inherent fixed-length generation paradigm of D-LLMs, generated sequences often contain many structural padding tokens, especially when the final answer is short. These tokens are introduced to maintain a fixed sequence length and usually carry little hallucination-related information.

To extract reliable uncertainty evidence, we employ a conservative filtering mechanism that only excludes task-invariant structural tokens from evidence calculation. Specifically, we define a structural token set $\mathcal{I}_{\text{struct}}$, including padding tokens, boundary markers, and empty positions, such as \texttt{<|endoftext|>} and \texttt{[PAD]}.

At each denoising step $t$, for token $x^{(t)}_i$ at position $i \in \{1, \dots, l\}$, we extract the subset of structurally valid positions $\mathcal{K}_t$:
\begin{equation}
    \mathcal{K}_t = \{ i \mid 1 \le i \le l, \, x^{(t)}_i \notin \mathcal{I}_{\text{struct}} \}.
\end{equation}
Confining evidence construction to $\mathcal{K}_t$ reduces structural noise caused by fixed-length generation while preserving potentially meaningful lexical and formatting tokens.

\paragraph{Statistical Evidence Construction.}
After removing structural padding tokens, the remaining tokens in $\mathcal{K}_t$ still contribute differentially to hallucination detection (e.g., factual nouns versus prepositions or pronouns). Consequently, a simple average of these tokens may smooth out localized signals and produce weak evidence. To better capture informative patterns, we adopt a statistics-based method to condense the filtered uncertainty field $\{s_{t,i}\}_{i \in \mathcal{K}_t}$ at each step $t$ into a statistical evidence vector $\mathbf{a}_t \in \mathbb{R}^3$. This captures the mean, peak, and top-$k$ characteristics of the entropy distribution:
\begin{equation}
    \mathbf{a}_t = \big[ \mu_{\text{avg}}^{(t)}, \, s_{\max}^{(t)}, \, \mu_{\text{top-}k}^{(t)} \big].
\end{equation}

These three statistical moments capture the filtered uncertainty at different granularities:
\ding{182}~\textbf{Global Uncertainty} ($\mu_{\text{avg}}^{(t)}$) tracks the mean uncertainty across $\mathcal{K}_t$, indicating macro-level convergence to a coherent state; 
\ding{183}~\textbf{Peak Uncertainty} ($s_{\max}^{(t)}$) captures the maximum entropy spike, revealing micro-level factual errors (e.g., incorrect entities) that are typically masked by global averaging; 
\ding{184}~\textbf{Regional Uncertainty} ($\mu_{\text{top-}k}^{(t)}$) averages the top-$k$ entropy (or all, if $|\mathcal{K}_t| < k$), capturing phrase-level uncertainty patterns among the remaining tokens. 
Together, these statistics allow the detector to capture errors at different scales and remain robust to varying sequence lengths. This approach ensures that isolated factual mistakes are not hidden within fluent and confident contexts.

Finally, stacking these step-wise statistics transforms the raw trajectory $\mathcal{T}$ into a compact evidence trajectory $\mathcal{E} = (\mathbf{a}_T, \mathbf{a}_{T-1}, \dots, \mathbf{a}_0)$, mapping the semantic uncertainty resolution from initial noise ($t=T$) to the final generation ($t=0$).

\subsection{Dynamical Deviation Learning}\label{sec:traj_learning}
Once high-quality evidence is constructed, the next question is how to leverage it for hallucination detection. While the above module focuses on the spatial dimension (i.e., structural-token filtering) of evidence construction, the temporal dimension is also crucial, namely how to model the denoising dynamics of the evidence to identify hallucinations. Although the evolution of evidence can reveal useful patterns, the range and trend of entropy may vary significantly depending on the specific scenarios and questions, making it difficult to directly use the evidence for hallucination prediction. 
To mine discriminative signals from denoising dynamics of evidence, we propose a dynamical deviation learning module consisting of two components.

First, we build an \textit{evidence dynamics generator} to predict an expected reference trajectory conditioned on the input question, which captures the specific evolution patterns of evidence under different Q\&A scenarios. 
Meanwhile, a \textit{deviation-based hallucination detector} makes the final prediction by adaptively measuring the discrepancy between the actual evidence trajectory $\mathcal{E}$ and the predicted reference, supporting reliable hallucination detection. 

\paragraph{Reference Evidence Dynamics Generator.}
Since denoising dynamics vary significantly across different queries and scenarios, raw evidence is not a reliable direct indicator for hallucination detection. Rather than predicting hallucinations directly from evidence, we design a generator $g_{\theta}$ to model the expected evidence trajectory. Conditioned on the input query embedding $\mathbf{q}$ and the timestep embedding $\mathbf{e}_t$, $g_{\theta}$ learns the relationship between the task content and the uncertainty dynamics. This produces a reference trajectory that represents normal, factual behavior to assist the detection process.

For each step $t$, the generator predicts a reference evidence vector $\hat{\mathbf{a}}_t = g_{\theta}(\mathbf{q}, \mathbf{e}_t)$. We optimize the generator using only factual samples ($y=0$) via:
\begin{equation}
    \mathcal{L}_{\text{ref}} = \mathbb{E}_{y=0} \left[ \frac{1}{T} \sum_{t=0}^{T} \text{SmoothL1}(\mathbf{a}_t, \hat{\mathbf{a}}_t) \right].
\end{equation}

This approach allows the framework to model task-specific dynamics of factual sequences. In practice, we first pre-train $g_{\theta}$ using $\mathcal{L}_{\text{ref}}$ to establish a stable reference space for the subsequent detection stage.

\paragraph{Deviation-based Hallucination Detector.}
Based on the reference evidence trajectory, we use the deviation between the actual and reference curves as the direct criterion for hallucination detection. The detector can infer hallucinations by adaptively evaluating differences in curve trends. Specifically, we first construct a composite feature vector $\mathbf{x}_t = [\mathbf{a}_t;\; \hat{\mathbf{a}}_t;\; \Delta\mathbf{a}_t]$ at each step $t$ to model the dynamics. Here, $\mathbf{a}_t$ and $\hat{\mathbf{a}}_t$ represent the observed and reference states, while $\Delta\mathbf{a}_t = \mathbf{a}_{t-1} - \mathbf{a}_{t}$ captures the step-wise velocity toward the subsequent state. A shared MLP, $f_{\phi}$, then projects these vectors into hidden representations $\mathbf{z}_t$.

Since different steps contribute differently to detection, we employ a learnable weight $\omega_t$ to model their importance. A scoring network evaluates each time step by computing unnormalized scores \begin{equation}
    u_t = \mathbf{w}^\top \tanh(\mathbf{U} \mathbf{z}_t + \mathbf{b}_u),
\end{equation}

\noindent where $\mathbf{w}$, $\mathbf{U}$, and $\mathbf{b}_u$ are learnable parameters. Then, these scores are normalized via a softmax function to produce the weight distribution $\omega_t = \exp(u_t) / \sum_{j=0}^T \exp(u_j)$. Finally, the hallucination prediction $\tilde{y}$ is computed by $\tilde{y} = \operatorname{MLP}\left(\sum_{t=0}^{T} \omega_t \mathbf{z}_t\right)$ , where temporal phases with prominent deviations can be prioritized.

\paragraph{Regularization Losses.}
To better identify hallucination characteristics, we introduce two dynamical regularizers weighted by $\omega_t$ that focus on late-stage stagnation and uncertainty rebound. Concretely, the path-deviation score $s_{\text{path}}$ quantifies cumulative divergence:

\begin{equation}
s_{\text{path}} = \sum_{t=0}^{T} \omega_t \sum_{d=1}^{D} |a_{t,d} - \hat{a}_{t,d}|.
\end{equation}

The rebound score $s_{\text{reb}}$ captures uncertainty increases along the descending steps ($T \!\to\! 0$):
\begin{equation}
s_{\text{reb}} = \sum_{t=1}^{T} \omega_t \sum_{d=1}^{D} \big(\max(0, a_{t-1,d} - a_{t,d})\big)^2.
\end{equation}

\newcolumntype{C}{>{\centering\arraybackslash}X}
\begin{table*}[t]
\centering
\small
\setlength{\tabcolsep}{3pt} 
\caption{AUROC(\%) comparison of hallucination detection methods on two D-LLMs across three QA datasets. The highest score is \textbf{bolded} and the second highest is \underline{underlined}.}
\label{tab:main}
\begin{tabularx}{\textwidth}{l l *{6}{C} c} 
\toprule
\multirow{2}{*}{\textbf{Model}} & \multirow{2}{*}{\textbf{Method}}
& \multicolumn{2}{c}{\textbf{TriviaQA}}
& \multicolumn{2}{c}{\textbf{HotpotQA}}
& \multicolumn{2}{c}{\textbf{CSQA}}
& \multirow{2}{*}{\textbf{Avg}} \\
\cmidrule(lr){3-4}\cmidrule(lr){5-6}\cmidrule(lr){7-8}
& & 128 & 64 & 128 & 64 & 128 & 64 & \\
\midrule

\rowcolor[HTML]{DEDEDE} \multicolumn{9}{l}{\textbf{LLaDA-8B-Instruct}} \\
\rowcolor[HTML]{F5F5F5} \multicolumn{9}{l}{\textit{Output-based Methods}} \\
& \quad Perplexity         & 50.4 & 47.6 & 49.3 & 51.2 & 65.6 & 65.0 & 54.9 \\
& \quad LN-Entropy         & 54.6 & 53.5 & 54.8 & 54.7 & 64.6 & 64.4 & 57.8 \\
& \quad Semantic Entropy   & 68.9 & 67.3 & 57.6 & 53.8 & 44.1 & 43.9 & 55.9 \\
& \quad Lexical Similarity & 62.5 & 59.0 & 64.2 & 57.1 & 57.3 & 60.7 & 60.1 \\
\rowcolor[HTML]{F5F5F5} \multicolumn{9}{l}{\textit{Latent-based Methods}} \\
& \quad EigenScore         & 69.2 & 66.9 & 64.7 & 59.2 & 58.5 & 60.6 & 63.2 \\
& \quad CCS                & 57.1 & 54.2 & 57.6 & 55.8 & 50.5 & 58.5 & 55.6 \\
& \quad TSV                & 60.2 & 61.1 & 65.0 & 59.4 & 52.9 & 55.2 & 59.0 \\
\rowcolor[HTML]{F5F5F5} \multicolumn{9}{l}{\textit{Trajectory-based Methods}} \\
& \quad TraceDet           & \underline{73.9} & \underline{74.1} & \underline{66.1} & \underline{63.7} & \underline{77.2} & \underline{77.1} & \underline{72.0} \\
& \quad \textbf{\ourmethod} & \textbf{86.6} & \textbf{86.2} & \textbf{84.3} & \textbf{85.2} & \textbf{81.5} & \textbf{81.2} & \textbf{84.2} \\

\midrule

\rowcolor[HTML]{DEDEDE} \multicolumn{9}{l}{\textbf{Dream-7B-Instruct}} \\
\rowcolor[HTML]{F5F5F5} \multicolumn{9}{l}{\textit{Output-based Methods}} \\
& \quad Semantic Entropy   & 73.7 & 72.5 & 62.7 & 67.7 & 51.4 & 48.6 & 62.8 \\
& \quad Lexical Similarity & 58.3 & 64.0 & 59.7 & 62.7 & 77.3 & 76.9 & 66.5 \\
\rowcolor[HTML]{F5F5F5} \multicolumn{9}{l}{\textit{Latent-based Methods}} \\
& \quad EigenScore         & 66.0 & 69.1 & 62.5 & 67.0 & 76.9 & 77.5 & 69.8 \\
& \quad CCS                & 56.9 & 50.3 & 51.7 & 58.2 & 54.2 & 53.2 & 54.1 \\
& \quad TSV                & \underline{75.6} & 74.7 & 58.7 & 63.0 & 62.3 & 56.8 & 65.2 \\
\rowcolor[HTML]{F5F5F5} \multicolumn{9}{l}{\textit{Trajectory-based Methods}} \\
& \quad TraceDet           & 78.1 & \textbf{86.7} & \underline{75.1} & \underline{76.0} & \textbf{84.7} & \underline{84.1} & \underline{80.8} \\
& \quad \textbf{\ourmethod} & \textbf{87.2} & \underline{84.5} & \textbf{80.2} & \textbf{85.4} & \underline{83.4} & \textbf{84.6} & \textbf{84.3} \\

\bottomrule
\end{tabularx}
\end{table*}

We define an adaptive margin $\hat{m}^{(\tau)}$ to represent the boundary of factual behaviors, updated via EMA using factual samples:
\begin{equation}
\begin{cases}
m_{\text{batch}} = \text{Quantile}(\{s_i \mid y_i=0\}, q) \\
\hat{m}^{(\tau)} = (1-\beta)\hat{m}^{(\tau-1)} + \beta m_{\text{batch}}.
\end{cases}
\end{equation}

The regularization losses are formulated as:
\begin{equation}
\mathcal{L}_{\text{path}} = \mathcal{L}_{\text{hinge}}(s_{\text{path}}; \hat{m}_{\text{path}}^{(\tau)}),
\end{equation}
\begin{equation}
\mathcal{L}_{\text{reb}} = \mathcal{L}_{\text{hinge}}(s_{\text{reb}}; \hat{m}_{\text{reb}}^{(\tau)}),
\end{equation}
where $\mathcal{L}_{\text{hinge}}(s; \hat{m}) = (1-y) \cdot s + y \cdot \max(0, \hat{m} - s)$~\cite{cortes-vapnik-1995-svm}. This forces the model to minimize deviations for factual samples while pushing hallucinations beyond the boundary. The total training objective combines the classification loss $\mathcal{L}_{\text{cls}}$ with these regularizers:
\begin{equation}
\mathcal{L} = \mathcal{L}_{\text{cls}} + \lambda_1 \mathcal{L}_{\text{path}} + \lambda_2 \mathcal{L}_{\text{reb}},
\end{equation}

\noindent where $\lambda_1$ and $\lambda_2$ are balance hyperparameters. Detailed pseudocode and complexity analysis of \ourmethod are provided in Appendix \ref{sec:appendix_algo}.

\section{Experiments}
\subsection{Experimental Setup}

\paragraph{Datasets and Models.}
To ensure a fair and rigorous comparison, we strictly adhere to the evaluation protocol established by TraceDet~\cite{chang-etal-2025-tracedet}. 
Specifically, our experiments are conducted across three widely adopted fact-based QA benchmarks:
TriviaQA~\cite{joshi-etal-2017-triviaqa}, HotpotQA~\cite{yang-etal-2018-hotpotqa}, and CommonsenseQA~\cite{talmor-etal-2019-commonsenseqa}, which cover open-domain fact-based questions,
multi-hop reasoning, and commonsense reasoning.
We adopt two representative models, LLaDA-8B-Instruct~\cite{nie-etal-2025-llada} and Dream-7B-Instruct~\cite{ye-etal-2025-dream}, as backbone D-LLMs. 

\begin{table}[t]
\centering
\small
\setlength{\tabcolsep}{2pt}
\caption{Zero-shot cross-task generalization matrix.} 
\label{tab:cross_transfer_hierarchical}

\resizebox{\columnwidth}{!}{
    \begin{tabularx}{1.1\columnwidth}{l *{6}{C} c}
    \toprule
    \multirow{2}{*}{Method}
    & \multicolumn{2}{c}{TriviaQA}
    & \multicolumn{2}{c}{HotpotQA}
    & \multicolumn{2}{c}{CSQA}
    & \multirow{2}{*}{Avg} \\
    \cmidrule(lr){2-3}\cmidrule(lr){4-5}\cmidrule(lr){6-7}
    & H-QA & CSQA & T-QA & CSQA & T-QA & H-QA & \\
    \midrule

    \rowcolor[HTML]{F5F5F5} \multicolumn{8}{c}{\textit{Latent-based}} \\
    CCS  & 50.1 & 54.5 & 51.8 & 54.0 & 54.2 & 56.6 & 53.5 \\
    TSV  & 58.5 & 65.3 & 65.5 & 59.1 & 56.2 & 63.2 & 61.3 \\
    \cmidrule(lr){1-8}
    
    \rowcolor[HTML]{F5F5F5} \multicolumn{8}{c}{\textit{Trajectory-based}} \\
    TraceDet & 73.1 & 61.5 & 57.4 & 65.0 & \textbf{74.8} & 66.2 & 66.3 \\
    \textbf{\ourmethod} & \textbf{85.4} & \textbf{68.6} & \textbf{73.3} & \textbf{65.4} & 73.6 & \textbf{71.2} & \textbf{72.9} \\

    \bottomrule
    \end{tabularx}
}
\end{table}

\paragraph{Baselines.}
We compare our method against the same set of baselines as TraceDet, covering both output-based and
latent-based hallucination detection approaches.
Output-based baselines include Perplexity~\cite{ren-etal-2022-out}, Length-Normalized Entropy (LN-Entropy)~\cite{malinin-gales-2020-uncertainty},
Semantic Entropy~\cite{kuhn-etal-2023-semantic}, and Lexical Similarity~\cite{lin-etal-2024-generating}.
Latent-based baselines include EigenScore~\cite{chen-etal-2024-inside}, Contrast-Consistent Search (CCS)~\cite{burns-etal-2022-discovering},
and Truthfulness Separator Vector (TSV)~\cite{park-etal-2025-steer}.
We also compare our framework against TraceDet~\cite{chang-etal-2025-tracedet}, the state-of-the-art method for D-LLM hallucination detection.

\begin{figure}[t]
    \centering
    \includegraphics[width=0.95\linewidth]{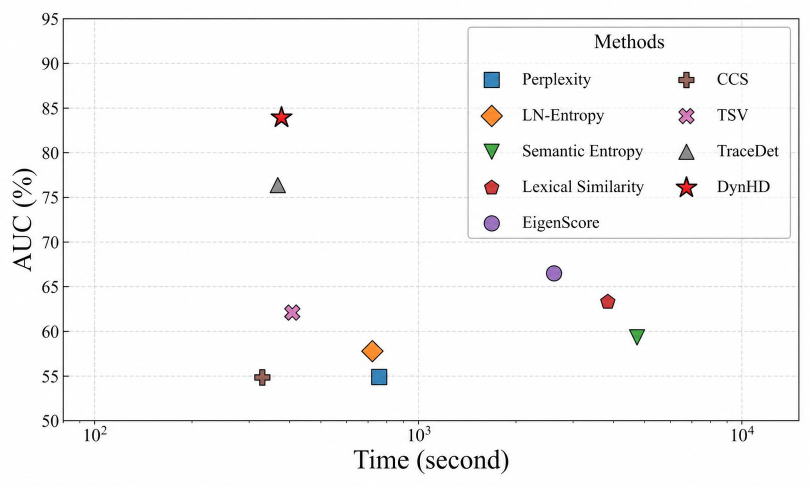}
    \caption{Pareto-frontier of performance vs. efficiency.} 
    \label{fig:tradeoff}
\end{figure}

\paragraph{Evaluation Protocol.}
We evaluate hallucination detection performance using AUROC, following prior work~\cite{park-etal-2025-steer,chen-etal-2024-inside}. 
Ground-truth correctness labels are provided by GPT-4o-mini, which is instructed to judge whether
a generated answer is factually correct based on its consistency with the reference answer and its own internal knowledge ~\cite{zheng-etal-2023-judging, liu-etal-2023-geval}.
We further assess the reliability of the automatic annotation by measuring agreement with human judgments, achieving an agreement rate of 94\%. 
We partition the dataset into 1,400 samples for training, 300 for validation, and 300 for testing. 
Model selection is based on validation AUROC, and all reported results are obtained on the held-out test set. 
We provide detailed descriptions of the baselines, benchmarks, and experimental settings in Appendix~\ref{sec:appendix_exp_detail}, with additional experiments in Appendix~\ref{sec:appendix_results}.

\begin{figure*}[t] 
    \centering
    \begin{subfigure}[b]{0.31\textwidth}
        \centering
        \includegraphics[width=\textwidth]{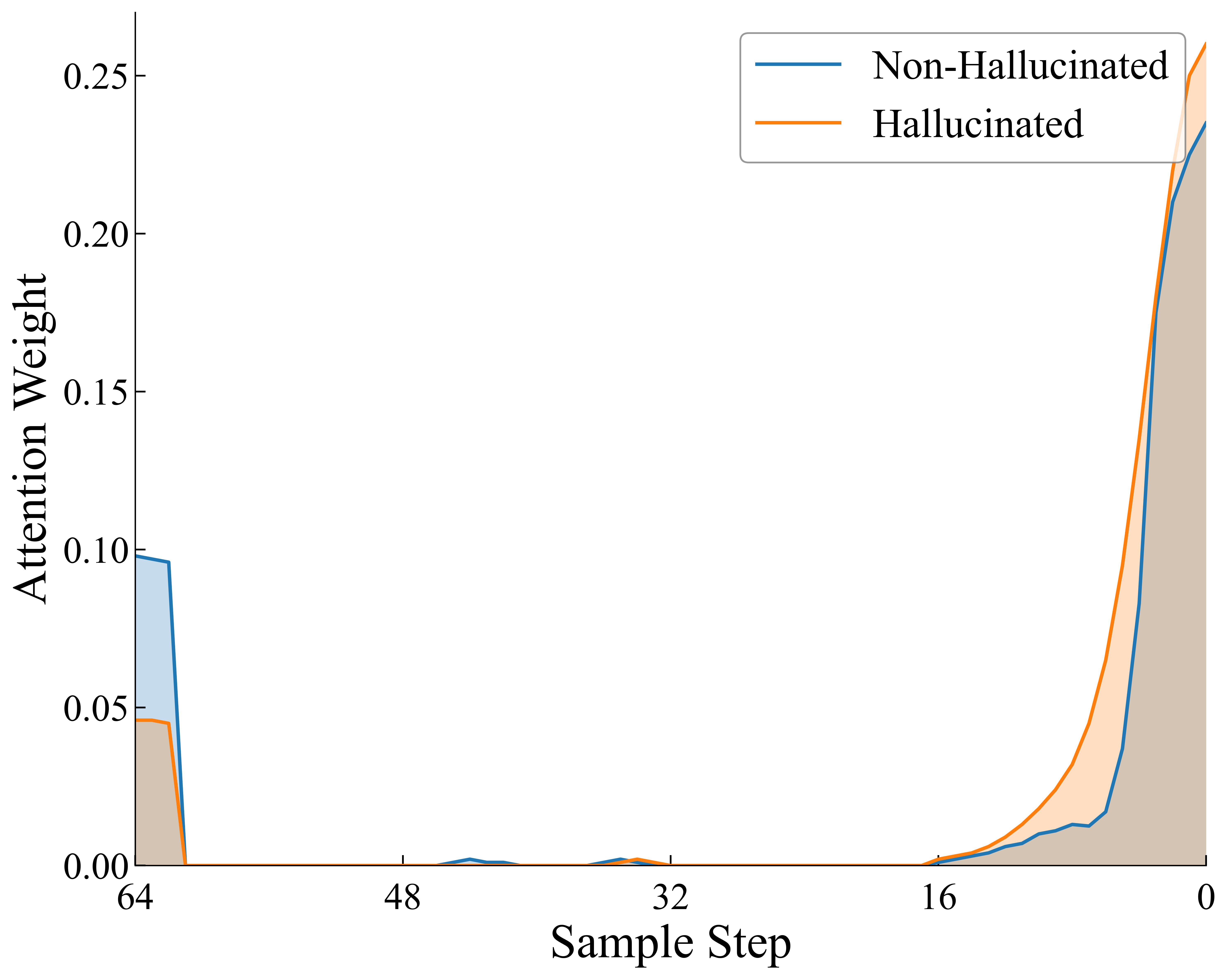}
        \caption{}
        \label{fig:attn_dist}
    \end{subfigure}%
    \hfill
    \begin{subfigure}[b]{0.31\textwidth}
        \centering
        \includegraphics[width=\textwidth]{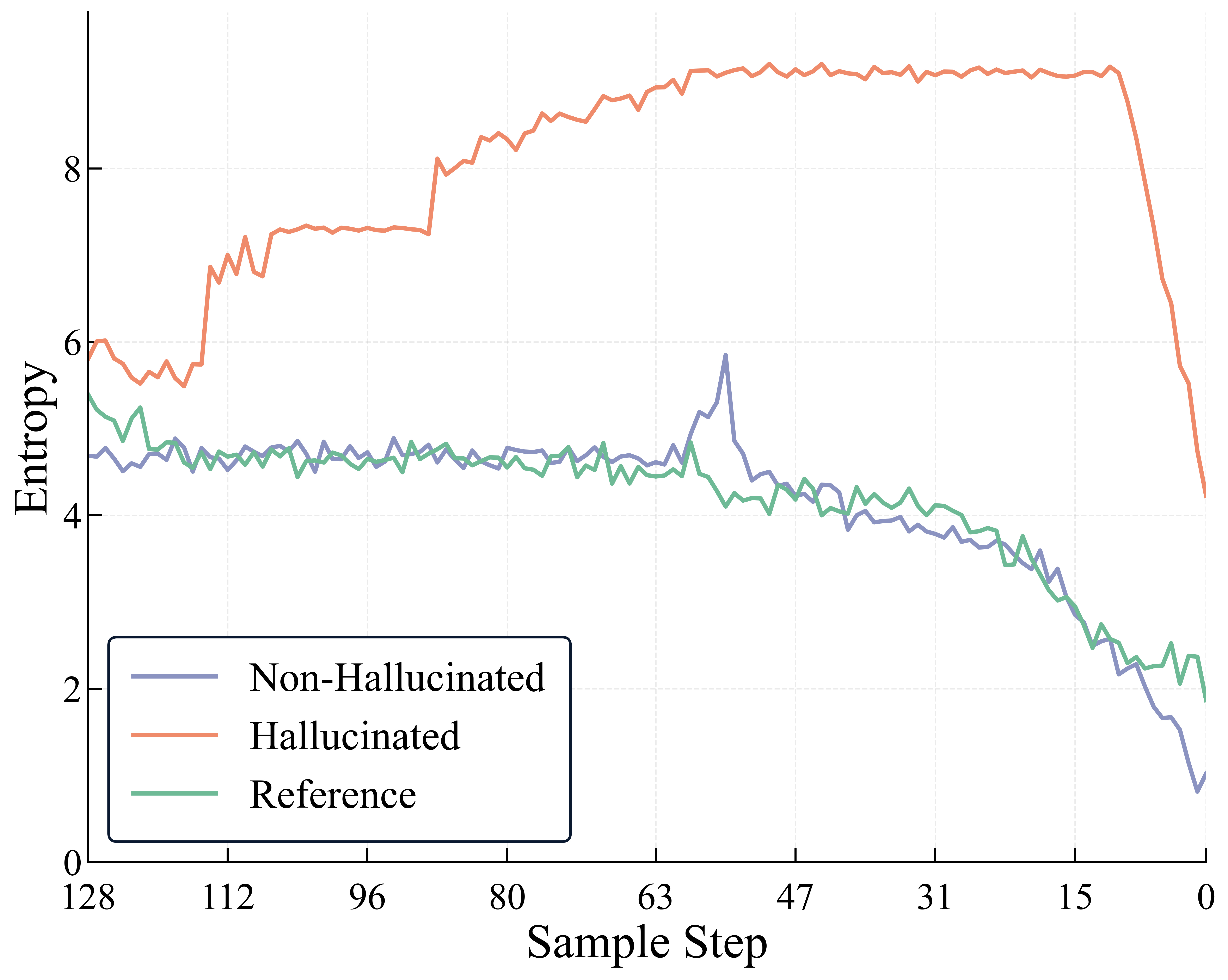}
        \caption{}
        \label{fig:trajectory_128}
    \end{subfigure}%
    \hfill
    \begin{subfigure}[b]{0.31\textwidth}
        \centering
        \includegraphics[width=\textwidth]{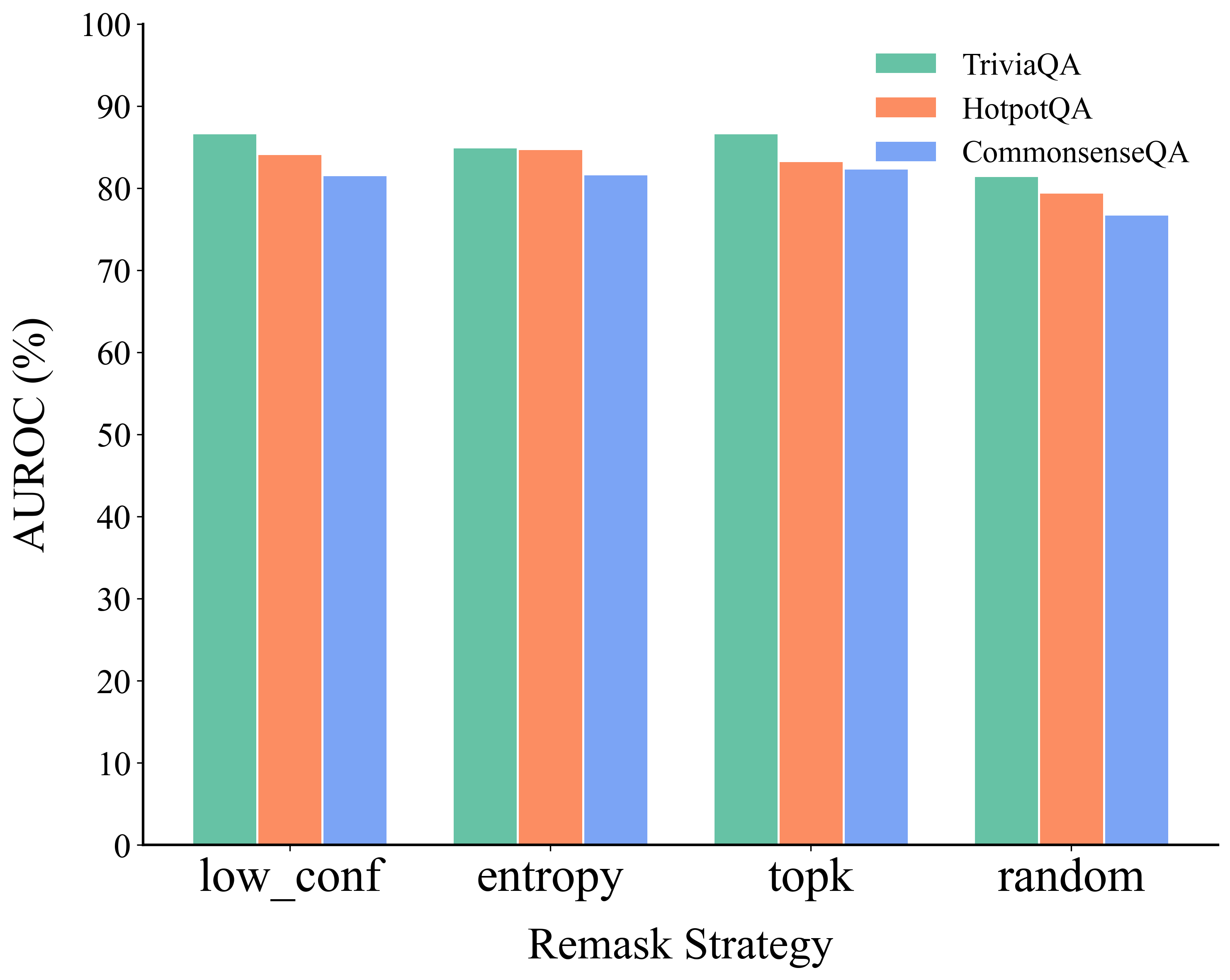}
        \caption{}
        \label{fig:masking_strategies}
    \end{subfigure}
    
    \caption{(a) Visualization of attention weights $\omega_t$ on different steps. (b) Top-$k$ entropy trajectories of representative samples on HotpotQA ($T=128$). (c) Performance comparison of different D-LLM remasking strategies.}
    \label{fig:mechanism_analysis}
\end{figure*}

\begin{table}[t]
\centering
\caption{Ablation study in AUROC(\%).} 
\label{tab:ablation_optimized_grey}
\resizebox{\columnwidth}{!}{%
\begin{tabular}{l cc cc cc}
\toprule
\multirow{2}{*}{\textbf{Ablation Variant}} & \multicolumn{2}{c}{\textbf{TriviaQA}} & \multicolumn{2}{c}{\textbf{HotpotQA}} & \multicolumn{2}{c}{\textbf{CSQA}} \\
\cmidrule(lr){2-3} \cmidrule(lr){4-5} \cmidrule(lr){6-7}
& 128 & 64 & 128 & 64 & 128 & 64 \\
\midrule
\rowcolor[HTML]{DEDEDE} \multicolumn{7}{l}{\textbf{LLaDA-8B-Instruct}} \\
\textbf{\ourmethod} (Full Framework) & \textbf{86.6} & \textbf{86.2} & \textbf{84.3} & \textbf{85.2} & \underline{81.5} & \textbf{84.2} \\
\midrule
\rowcolor[HTML]{F5F5F5} \multicolumn{7}{l}{\textbf{Semantic-aware Evidence Construction Components}} \\
\quad w/o Token Filtering & 74.8 & 73.6 & 75.5 & 78.1 & 74.1 & 76.6 \\
\quad w/o Mean Uncertainty ($\mu_{\text{avg}}$) & 78.6 & 81.1 & 79.0 & 80.2 & 78.5 & 77.1 \\
\quad w/o Max Entropy ($s_{\max}$) & 82.5 & 83.1 & 81.8 & 82.3 & 79.1 & 82.1 \\
\quad w/o Top-$k$ Entropy ($\mu_{\text{top-}k}$) & 82.7 & 82.6 & 81.7 & 83.0 & 78.9 & 82.5 \\
\midrule
\rowcolor[HTML]{F5F5F5} \multicolumn{7}{l}{\textbf{Dynamical Deviation Learning Components}} \\
\quad Naive Average  & 57.3 & 65.5 & 51.7 & 48.3 & 63.2 & 61.8 \\
\quad Flattened Temporal Aggregation  & 62.3 & 70.5 & 66.8 & 57.5 & 67.1 & 66.2 \\
\quad Uniform Temporal Pooling (w/o Attn) & 84.5 & 84.8 & 82.4 & 83.1 & 79.5 & 80.5 \\
\quad w/o $\mathcal{L}_{path}$ (Stagnation) & 84.1 & 85.3 & 84.1 & \underline{83.8} & \textbf{81.6} & 83.2 \\
\quad w/o $\mathcal{L}_{reb}$ (Rebound) & \underline{85.2} & \underline{85.6} & \underline{84.2} & 83.5 & 80.6 & \underline{83.3} \\
\midrule
\rowcolor[HTML]{DEDEDE} \multicolumn{7}{l}{\textbf{Dream-7B-Instruct}} \\
\textbf{\ourmethod} (Full Framework) & \textbf{87.2} & \textbf{84.5} & \textbf{80.2} & \textbf{85.4} & \textbf{83.4} & \underline{84.6} \\
\midrule
\rowcolor[HTML]{F5F5F5} \multicolumn{7}{l}{\textbf{Semantic-aware Evidence Construction Components}} \\
\quad w/o Token Filtering & 75.4 & 72.1 & 69.8 & 74.2 & 73.6 & 73.9 \\
\quad w/o Mean Uncertainty ($\mu_{\text{avg}}$) & 79.2 & 78.5 & 74.5 & 79.1 & 76.2 & 77.8 \\
\quad w/o Max Entropy ($s_{\max}$) & 83.1 & 81.2 & 77.5 & 82.3 & 80.9 & 81.5 \\
\quad w/o Top-$k$ Entropy ($\mu_{\text{top-}k}$) & 83.5 & 80.9 & 77.8 & 82.6 & 80.5 & 81.9 \\
\midrule
\rowcolor[HTML]{F5F5F5} \multicolumn{7}{l}{\textbf{Dynamical Deviation Learning Components}} \\
\quad Naive Average  & 58.2 & 59.1 & 48.5 & 52.3 & 60.1 & 57.8 \\
\quad Flattened Temporal Aggregation  & 65.1 & 66.3 & 59.4 & 62.7 & 68.2 & 65.6 \\
\quad Uniform Temporal Pooling (w/o Attn) & 84.5 & 82.1 & 78.0 & 83.1 & 81.2 & 81.8 \\
\quad w/o $\mathcal{L}_{path}$ (Stagnation) & 85.2 & \underline{83.5} & 79.2 & \underline{84.5} & \underline{82.8} & 83.8 \\
\quad w/o $\mathcal{L}_{reb}$ (Rebound) & \underline{86.1} & 83.1 & \underline{79.5} & 84.2 & 82.7 & \textbf{84.8} \\
\bottomrule
\end{tabular}%
}
\end{table}

\subsection{Experimental Results}
\paragraph{Main Results.} 
Table~\ref{tab:main} presents a comprehensive comparison of \ourmethod against baseline methods across two D-LLMs and three factuality QA datasets with varying generation lengths. From the results, we have several observations. \ding{182}~\ourmethod achieves the highest performance in almost all settings. On the LLaDA-8B-Instruct model, it outperforms the strongest baseline by an average of 12.2\% absolute AUROC, while on Dream-7B-Instruct, it reaches an average of 84.3\%. These consistent gains exhibit the value of exploiting fine-grained deviation dynamics and convergence morphology rather than relying solely on output uncertainty or static hidden-state representations. \ding{183}~Output-based methods lack consistency across datasets, showing poor performance on several cases. This occurs because multiple-choice answers are often confident but wrong, rendering uncertainty-based metrics ineffective. \ding{184}~Latent-based approaches like EigenScore and TSV show more competitive results on Dream-7B-Instruct by capturing hidden-state geometry, but remain sensitive to dataset shifts.

\paragraph{Inference Efficiency.} 
Figure~\ref{fig:tradeoff} shows the balance between performance and efficiency. \ourmethod occupies the upper-left quadrant, representing the best results in our evaluation. While its speed is comparable to fast baselines like CCS, \ourmethod achieves significantly higher accuracy. Notably, our approach is much more efficient than multi-sample methods like Semantic Entropy, which often have high computational costs caused by multi-round sampling. Overall, \ourmethod provides an ideal balance of accuracy and speed, allowing for real-time monitoring without causing extra delays.

\paragraph{Cross-dataset Generalization Ability.}
To evaluate robustness, we conduct zero-shot cross-dataset assessments in Table~\ref{tab:cross_transfer_hierarchical}, where \ourmethod yields the highest average AUROC (72.9\%). It notably achieves an 85.5\% AUROC in TriviaQA to H-QA transfer, significantly outperforming TraceDet (73.1\%) and suggesting that our dynamical encoding captures universal denoising laws across domains. On the challenging CSQA source, \ourmethod remains robust (e.g., 73.6\% on T-QA), despite slightly underperforming TraceDet. This is because the fixed options lead to a faster decision process, which reduces the observable uncertainty signals. 

\paragraph{Ablation Study.} 
As summarized in Table~\ref{tab:ablation_optimized_grey}, we evaluate the contributions of key designs in \ourmethod, coming with the following findings. 
 
In terms of the evidence construction module, \ding{182}~without token filtering, the performance drops significantly, which proves that filtering out noise is essential to get clean signals. \ding{183}~Removing individual statistical evidence components consistently degrades results, proving that robust detection requires simultaneous monitoring of semantic stability, localized anomaly, and regional instability. 
 
For the deviation learning module, \ding{184}~ \textit{Naive Average} variant yields poor performance because simple arithmetic averaging cannot effectively distinguish hallucinations. \ding{185}~\textit{Flattened Temporal Aggregation} underperforms due to its lack of explicit temporal dynamic analysis. \ding{186}~Removing attention-based weighting degrades performance, as the model fails to capture critical dynamic changes, especially at the late stages of the denoising process. \ding{187}~Our regularization terms further improve the model's ability to differentiate hallucinations from factual generations by capturing and penalizing specific non-convergent trajectories, such as \textit{Stagnation} and \textit{Rebound} patterns.

\paragraph{Visualization of Temporal Weighting.} 
Figure~\ref{fig:attn_dist} shows the attention weights for LLaDA-8B-Instruct on the TriviaQA dataset ($t=64$). We can see the learned weights are mostly concentrated in the final steps of the denoising process. This matches our observation that stagnation and rebound patterns usually happen at the end of the trajectory. By focusing on these late steps, our model can find signs of hallucinations and ignore early random noise.

\paragraph{Visualization of Reference Trajectories.} 
Figure~\ref{fig:trajectory_128} shows entropy trajectories on the HotpotQA dataset. We introduce the learned reference path to show how a correct denoising process should look. The factual sample stays close to this path and drops smoothly toward zero. In contrast, a hallucinated sample is unstable and shows a clear rebound at the end. This shows that the curve's shape matters more than its absolute value. The reference generator facilitates identifying denoising processes associated with hallucinations.

\paragraph{Robustness to Masking Strategies.} 
Figure~\ref{fig:masking_strategies} examines the performance of \ourmethod on LLaDA-8B-Instruct under various internal remasking strategies employed by D-LLMs during generation. The results across multiple datasets consistently demonstrate competitive detection performance, regardless of the specific spatial filtering logic used (such as entropy-based or confidence-based remasking). This stability suggests that the proposed framework is adaptable and maintains its effectiveness across diverse D-LLM configurations.

\section{Conclusion}
We propose \ourmethod to detect D-LLM hallucinations using trajectory-based deviation modeling. \ourmethod identifies factual errors by comparing actual evidence against a learned reference. This approach captures key signals such as late-stage stagnation and uncertainty rebound. By combining semantic-aware evidence construction with dynamical deviation learning, \ourmethod provides a practical framework for exploiting both spatial uncertainty patterns and temporal denoising behaviors. Experiments on mainstream models show that \ourmethod achieves superior performance with high computational efficiency. Overall, this work demonstrates that denoising dynamics can serve as signals for hallucination detection of D-LLMs.

\section*{Acknowledgment}
The work of Y. Liu was partially supported by the Australian Research Council (ARC) under Grant No. DE260101172.

\section*{Limitations}
Although \ourmethod{} demonstrates strong performance on standard QA benchmarks, the current evaluation mainly focuses on factual question answering scenarios, including open-domain QA, multi-hop QA, and commonsense QA. These benchmarks provide clear correctness criteria and are suitable for controlled hallucination detection evaluation. Nevertheless, hallucination detection in more open-ended generation settings, such as long-form summarization, multi-turn dialogue, mathematical reasoning, code generation, and structured prediction, may involve more diverse factual boundaries and formatting patterns. Evaluating \ourmethod{} in these scenarios would further strengthen the understanding of its generality.
In addition, our experiments are conducted on two representative D-LLM backbones under commonly used generation lengths. While the results demonstrate consistent effectiveness across different datasets and models, future work could further examine broader model families, decoding configurations, and multilingual settings. Finally, the current design relies on training an evidence dynamics generator using in-domain factual samples to establish reference trajectories, which may affect its adaptability to entirely unseen domains. Since our framework is tailored to the continuous denoising process of Diffusion LLMs, extending these trajectory-based insights to standard auto-regressive models remains an interesting direction for future research.

\section*{Ethics Statement}
Our research focuses exclusively on scientific questions, with no involvement of human subjects, animals, or environmentally sensitive materials. Therefore, we foresee no ethical risks or conflicts of interest. We are committed to maintaining the highest standards of scientific integrity and ethics to ensure the validity and reliability of our findings.

\section*{Artifact Use and Licensing}
We use publicly available datasets, models, and baseline methods in this work, including TriviaQA, HotpotQA, CommonsenseQA, LLaDA-8B-Instruct, and Dream-7B-Instruct. 
We cite the original creators of these artifacts and use them only for research and evaluation purposes. 
We do not redistribute the original datasets or model weights, and our use follows the corresponding licenses and terms of use released by the artifact providers.

\bibliography{custom}

\clearpage

\appendix
\begin{algorithm*}[t]
\caption{The two-stage training algorithm of \ourmethod{}}
\label{alg:dynHD}
\begin{algorithmic}[1]
\REQUIRE Training pool $\mathcal{X}$; ignore set $\mathcal{I}_{\text{ignore}}$; hyperparameter $k$; epochs $E_1, E_2$; batch size $B$.
\ENSURE Optimized reference parameters $\theta$ and detector parameters $\Phi$.

\STATE // \textbf{Stage 1: Evidence Dynamics Generator (Sec. 3.2)}
\STATE Initialize generator parameters $\theta$;
\FOR{$epoch = 1$ \TO $E_1$}
    \STATE Sample mini-batch $\{(\mathcal{T}, \mathbf{q}, y)_i\}_{i=1}^B \subset \mathcal{X}$ where $y=0$ \COMMENT{Factual samples only}
    \FORALL{$(\mathcal{T}, \mathbf{q})$ \textbf{in} mini-batch}
        \STATE \textbf{Evidence Construction:} Extract valid semantic positions $\mathcal{K}_t$ via $\mathcal{I}_{\text{ignore}}$;
        \STATE Condense step-wise uncertainty into $\mathbf{a}_t = [\mu_{\text{keep}}^{(t)}, s_{\max}^{(t)}, \mu_{\text{top-}k}^{(t)}]$ (Eq. 4);
        \STATE Predict reference trajectory $\hat{\mathbf{a}}_t = g_\theta(\mathbf{q}, \mathbf{e}_t)$ using query and timestep embeddings;
    \ENDFOR
    \STATE Update $\theta$ by minimizing $\mathcal{L}_{\text{ref}}$ via SmoothL1 loss (Eq. 5);
\ENDFOR

\STATE // \textbf{Stage 2: Deviation-based Hallucination Detector (Sec. 3.2)}
\STATE Initialize detector parameters $\Phi$ and adaptive factual margins $\hat{m}^{(\tau)}$;
\FOR{$epoch = 1$ \TO $E_2$}
    \STATE Sample mini-batch $\{(\mathcal{T}, \mathbf{q}, y)_i\}_{i=1}^B \subset \mathcal{X}$;
    \FORALL{$(\mathcal{T}, \mathbf{q}, y)$ \textbf{in} mini-batch}
        \STATE \textbf{Evidence Construction:} Obtain actual evidence trajectory $\mathcal{E} = (\mathbf{a}_T, \dots, \mathbf{a}_0)$ from $\mathcal{T}$ via Eq. 1 \& 2;
        \STATE Generate reference trajectory $\hat{\mathcal{E}} = (\hat{\mathbf{a}}_T, \dots, \hat{\mathbf{a}}_0)$ via frozen $g_\theta$;
        \STATE Compute velocity $\Delta\mathbf{a}_t = \mathbf{a}_{t-1} - \mathbf{a}_{t}$ to capture transition dynamics;
        \STATE Construct composite feature $\mathbf{x}_t = [\mathbf{a}_t; \hat{\mathbf{a}}_t; \Delta\mathbf{a}_t]$ for each step $t$;
        \STATE Project $\mathbf{x}_t$ to $\mathbf{z}_t$, compute weight $\omega_t$ (Eq. 6), and aggregate to $\mathbf{h}$;
        \STATE Compute path-deviation score $s_{\text{path}}$ (Eq. 7) and rebound score $s_{\text{reb}}$ (Eq. 8);
    \ENDFOR
    \STATE Update adaptive margins $\hat{m}^{(\tau)}$ via EMA using scores of factual samples ($y=0$) in the batch (Eq. 9);
    \STATE Update $\Phi$ by minimizing total loss $\mathcal{L} = \mathcal{L}_{\text{cls}} + \lambda_1 \mathcal{L}_{\text{path}} + \lambda_2 \mathcal{L}_{\text{reb}}$ (Eq. 10);
\ENDFOR
\end{algorithmic}
\end{algorithm*}

\section{Related Work}
\label{sec:appendix_rw}

\paragraph{Hallucination Detection}
Most existing hallucination detection studies focus on Autoregressive Language Models (AR-LLMs)~\cite{zhang-etal-2025-siren,shen2026raising}. 
These methods typically extract signals from the final prediction, such as token-level uncertainty measures (e.g., entropy or perplexity~\cite{malinin-gales-2020-uncertainty,ren-etal-2022-out}), or assess semantic consistency across multiple samples (e.g., semantic entropy~\cite{kuhn-etal-2023-semantic, manakul-etal-2023-selfcheckgpt,li2026towards}). 
Some recent work also probes internal representations during the forward pass (e.g., hidden states~\cite{azaria-mitchell-2023-internal,li-etal-2023-inference,marks-tegmark-2023-geometry,liu2026rethinking}) to infer factuality. 
However, diffusion large language models (D-LLMs) differ fundamentally from AR-LLMs in their generation mechanism: hallucination cues are not confined to the final step but are distributed across a denoising trajectory spanning the entire iterative process. 
This process-centric nature makes static, outcome-based detectors prone to missing discriminative evidence exposed along the trajectory~\cite{chuang-etal-2023-dola}.

\paragraph{Diffusion Large Language Models}
Diffusion Large Language Models (D-LLMs) extend the remarkable success of diffusion paradigms~\cite{wang-etal-2025-diffusion} to the discrete text domain~\cite{li-etal-2022-diffusionlm}. 
By unifying the discrete remasking process, LLaDA has successfully scaled D-LLMs to the 8B parameter level, achieving performance parity with leading AR-LLMs such as LLaMA-3. 
Furthermore, Dream-7B demonstrates the architectural versatility of this paradigm by adopting the identical configurations of Qwen2.5-7B~\cite{qwen-2024-qwen25} and training it under a diffusion objective.
On the reliability side, \citet{wang-etal-2025-time} first identified the phenomenon of temporal oscillation, where correct answers may emerge at intermediate denoising steps but can be overridden by erroneous late-stage refinements. 
Building on the trajectory view, TraceDet~\cite{chang-etal-2025-tracedet} proposes a trajectory-based detection framework that models generation as an action trace and leverages the Information Bottleneck principle to automatically select the most discriminative sub-trace steps~\cite{paranjape2020information}. 
These studies motivate the use of denoising trajectories as process-level signals for D-LLM hallucination detection.

\paragraph{Distinction from TraceDet}
TraceDet~\cite{chang-etal-2025-tracedet} is the most relevant work to ours, as it also exploits the denoising trajectory of D-LLMs for hallucination detection. 
However, \ourmethod{} differs from TraceDet in two key aspects. 
First, TraceDet focuses on selecting discriminative sub-trace steps from the generation trajectory, but does not explicitly address the spatial information density imbalance caused by fixed-length generation. 
In contrast, \ourmethod{} introduces semantic-aware token filtering to remove structural padding tokens before evidence construction, reducing the dilution effect of non-semantic positions. 
Second, TraceDet mainly relies on selected trajectory fragments for detection, whereas \ourmethod{} models the continuous evolution of uncertainty evidence by learning a question-conditioned reference trajectory. 
Hallucinations are then detected by measuring deviations between the observed evidence trajectory and the learned reference dynamics. 
Therefore, \ourmethod{} extends trajectory-based hallucination detection from sub-trace selection to fine-grained spatial evidence construction and temporal deviation modeling.

\paragraph{Denoising Dynamics}
Denoising dynamics characterizes how a model progressively removes randomness and approaches the target distribution through multi-step refinement. 
\citet{li-cai-2025-breaking,li2026ofa} provide theoretical analysis suggesting that, under ideal conditions, the sampling error of D-LLMs should decrease as the number of denoising steps increases and eventually converge to the data distribution. 
This implies that reliable generation trajectories should be accompanied by smooth uncertainty dissipation. 
When hallucinations arise due to insufficient knowledge~\cite{yin-etal-2023-know,jiang2024large}, this dynamical balance can break: uncertainty may remain persistently high because of unresolved inconsistencies, leading to stagnation, or exhibit abnormal rebound in mid-to-late stages when refinement fails~\cite{huang2025survey}. 
Such morphological deviations from a normal convergence path directly reflect instability in the generation process~\cite{pavlova-wei-2025-lownoise,liu2026few}. 
This work is the first to utilize such continuous temporal evolution patterns for D-LLM hallucination detection, advancing the field from static uncertainty summaries toward a comprehensive process-level characterization of generative reliability.

\section{Algorithmic Details of \ourmethod}
\label{sec:appendix_algo}

\subsection{Pseudocode}
Algorithm~\ref{alg:dynHD} details the complete process of our algorithm.

\begin{figure*}[t]
    \centering
    \begin{subfigure}[b]{0.32\textwidth}
        \centering
        \includegraphics[width=\textwidth]{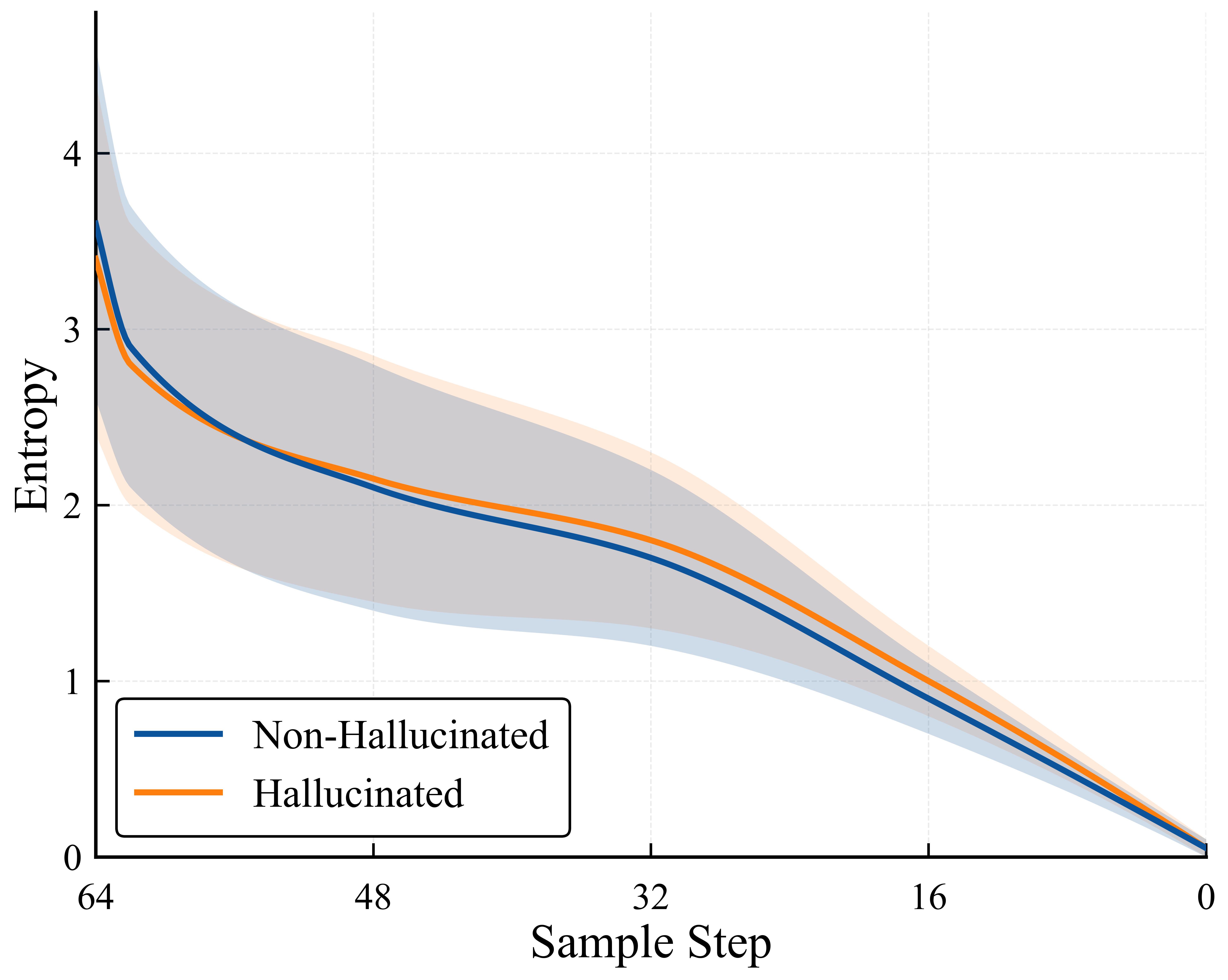}
        \caption{Global Distinguishability}
        \label{fig:global_dist}
    \end{subfigure}
    \hfill
    \begin{subfigure}[b]{0.32\textwidth}
        \centering
        \includegraphics[width=\textwidth]{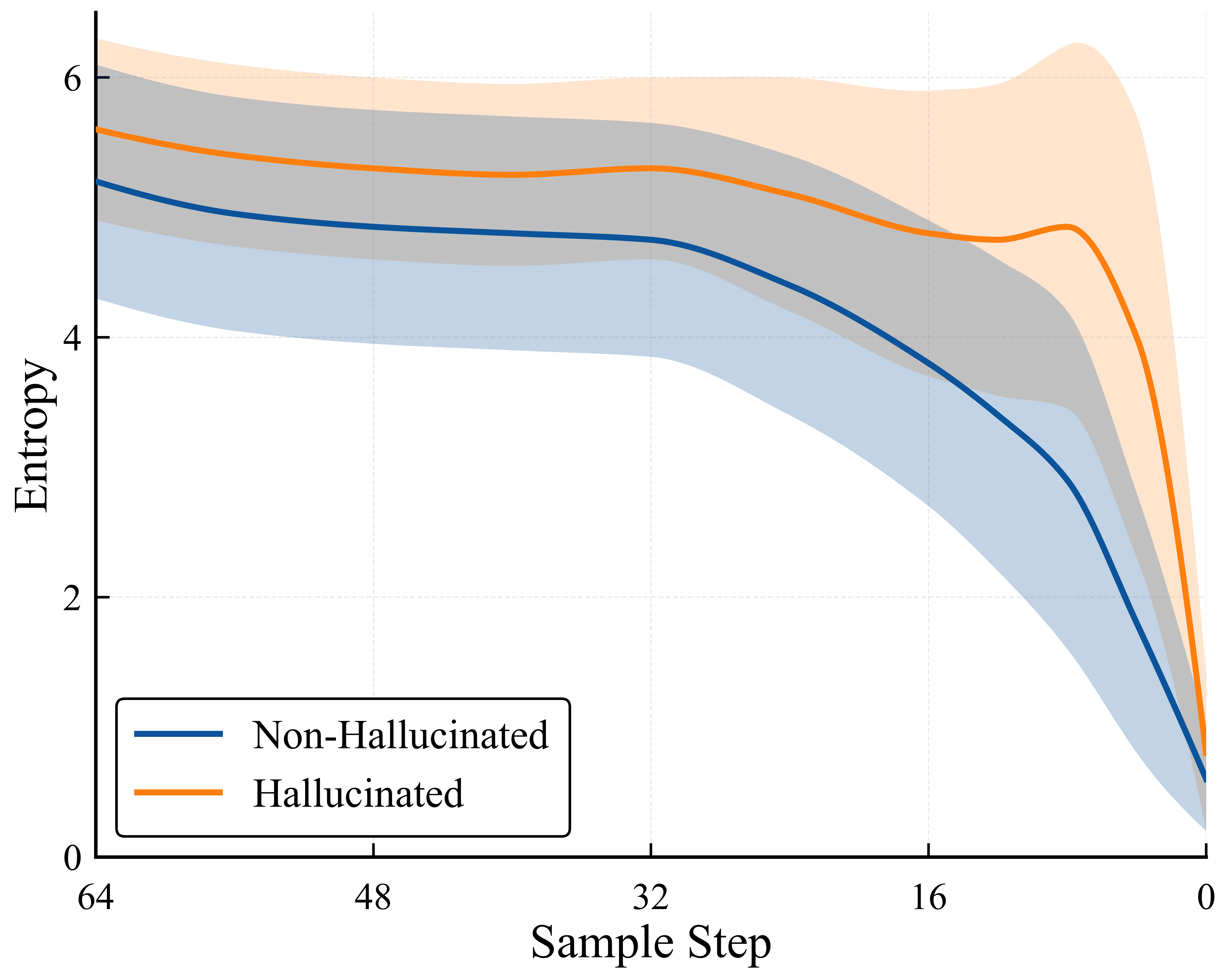}
        \caption{HotpotQA Results}
        \label{fig:hotpot_dist}
    \end{subfigure}
    \hfill
    \begin{subfigure}[b]{0.32\textwidth}
        \centering
        \includegraphics[width=\textwidth]{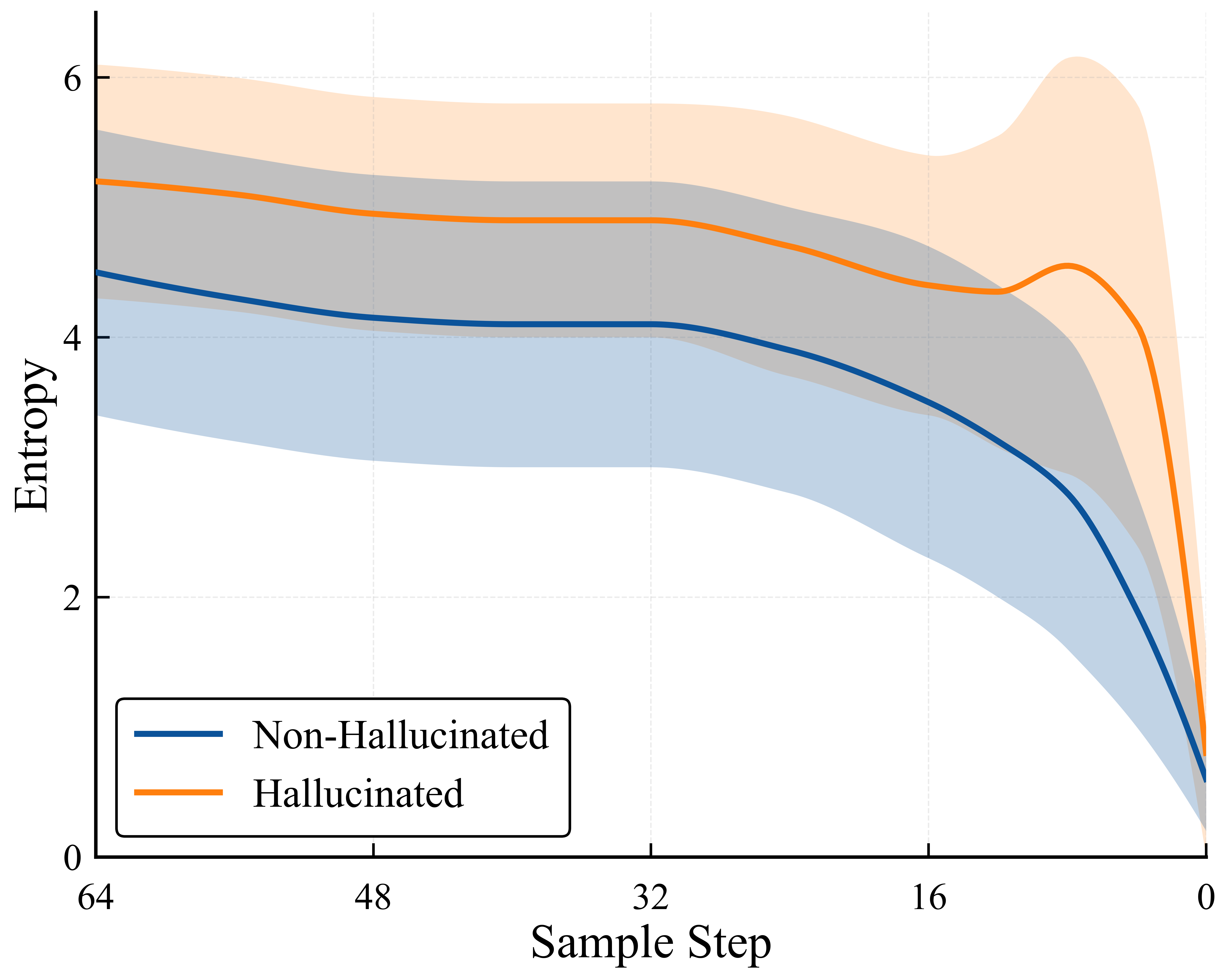}
        \caption{TriviaQA Results}
        \label{fig:trivia_dist}
    \end{subfigure}
    \caption{Empirical analysis of the distinguishability of statistical evidence trajectories. (a) shows the enhanced separation after structure-aware filtering, while (b) and (c) demonstrate the consistent morphological divergence between factual and hallucinatory samples across datasets.}
    \label{fig:evidence_distinguishability}
\end{figure*}

\subsection{Complexity Analysis}
\label{sec:complexity}
The computational overhead of \ourmethod is negligible compared to the D-LLM inference cost, as it operates on pre-computed uncertainty fields without re-invoking the backbone.

\begin{itemize}[leftmargin=*]
    \item \textbf{Spatial Filtering:} For each step $t$, position-wise masking is $\mathcal{O}(l)$, and extracting Top-$k$ statistics requires $\mathcal{O}(l \log k)$ using a min-heap. The total complexity for a full trajectory is $\mathcal{O}(T \cdot l \log k)$. Since the sequence length $l$ is typically much smaller than the vocabulary size $|\mathcal{V}|$, this process is significantly faster than the standard logit-to-probability conversion in Transformer heads.
    
    \item \textbf{Temporal Modeling:} Our temporal encoder and attention mechanism operate on the condensed 3D evidence vectors. With a hidden dimension $h$, the complexity is $\mathcal{O}(T \cdot (d \cdot h + h^2))$, where $d=3$. Since $d$ and $h$ are small constants independent of the LLM's scale, the inference latency of the detection head is essentially constant, adding no perceptible delay to the generation pipeline.
\end{itemize}

As a result, \ourmethod{} maintains an efficient $\mathcal{O}(T \cdot l)$ footprint, which is dwarfed by the quadratic self-attention complexity $\mathcal{O}(T \cdot l^2)$ of the underlying D-LLM.

\section{Detailed Experimental Setup}
\label{sec:appendix_exp_detail}

\subsection{Datasets and Benchmarks}
To ensure a rigorous and multi-faceted evaluation, we conduct experiments across three representative factuality-oriented QA benchmarks, following the protocol established in recent trajectory-based studies~\cite{chang-etal-2025-tracedet}:
\begin{itemize}[leftmargin=*]
    \item \textbf{TriviaQA}~\cite{joshi-etal-2017-triviaqa}: A large-scale open-domain dataset focusing on precise entity retrieval and factual knowledge.
    \item \textbf{HotpotQA}~\cite{yang-etal-2018-hotpotqa}: A multi-hop reasoning benchmark that requires synthesizing information across multiple context fragments to maintain logical consistency.
    \item \textbf{CommonsenseQA}~\cite{talmor-etal-2019-commonsenseqa}: A task evaluating general reasoning capabilities, where the model must navigate commonsense logic beyond surface-level pattern matching.
\end{itemize}
All dataset preprocessing, including prompt templates and data splits, strictly adheres to the configurations used in the comparative studies to ensure a level playing field.

\subsection{Baseline Configurations}
We compare \ourmethod{} against seven established baselines, categorized into two primary paradigms based on their information source:

\textbf{(1) Output-Centric Approaches:} These methods detect hallucinations by analyzing the generated text or its output probability distribution:
\begin{itemize}[leftmargin=*]
    \item \textit{Perplexity}~\cite{ren-etal-2022-out}: Identifies potential hallucinations based on the negative log-likelihood assigned by the model to its own sequence.
    \item \textit{LN-Entropy}~\cite{malinin-gales-2020-uncertainty}: Employs length-normalized predictive entropy to flag generations with unusually high average uncertainty.
    \item \textit{Semantic Entropy}~\cite{kuhn-etal-2023-semantic}: Quantifies consistency by partitioning multiple stochastic outputs into semantic equivalence classes.
    \item \textit{Lexical Similarity}~\cite{lin-etal-2024-generating}: Assesses hallucinations via surface-level lexical overlap across independent generation paths.
\end{itemize}

\textbf{(2) Latent-Centric Approaches:} These methods probe the internal representations and hidden states of the transformer architecture:
\begin{itemize}[leftmargin=*]
    \item \textit{EigenScore}~\cite{chen-etal-2024-inside}: Evaluates response consistency through the covariance structure of internal latent embeddings.
    \item \textit{CCS}~\cite{burns-etal-2022-discovering}: Searches for a contrastive-consistent direction in the latent space to distinguish truthful from untruthful activations.
    \item \textit{TSV}~\cite{park-etal-2025-steer}: Utilizes a separator vector to classify truthfulness based on learned centroids within the latent space.
\end{itemize}

Finally, we include \textbf{TraceDet}~\cite{chang-etal-2025-tracedet} as the primary trajectory-based baseline. TraceDet characterizes the denoising path using a uniform temporal aggregation strategy, providing a representative point of comparison for analyzing the evolution of uncertainty during the denoising process.

\subsection{Implementation Setup}
\label{sec:implementation}

Our experiments, including model inference and detector training, are conducted on \textbf{NVIDIA RTX 3090 GPUs (24GB VRAM)}. 

\subsection{Automatic Labeling Reliability}
Following recent hallucination detection studies, we use GPT-4o-mini to obtain correctness labels for generated answers. 
The judge is instructed to determine whether a generated answer is factually correct based on its consistency with the reference answer and its own knowledge. 
This reference-grounded protocol provides scalable and consistent annotations for both training and evaluation.

We acknowledge that automatic judges may introduce bias, especially for ambiguous cases or semantically equivalent answers. 
To assess the reliability of the annotations, we manually examine a subset of labeled samples and observe a 94\% agreement between GPT-4o-mini and human judgments. 
This agreement is also higher than the baseline agreement level used in our reliability check, further supporting the quality of the automatic annotations. 
This suggests that the automatic labels are sufficiently reliable for our controlled QA evaluation setting, while future work can further strengthen the annotation process with larger-scale human verification or multi-judge consensus.

\subsection{Hyperparameter Search Space}
\label{sec:hyperparams}

We conduct a grid search to optimize the performance of \ourmethod{}. The variables and their corresponding search ranges are summarized in Table~\ref{tab:hyperparams_compact}.

\begin{table}[ht]
\centering
\caption{Search space of key hyperparameters in \ourmethod{}.}
\label{tab:hyperparams_compact}
\resizebox{0.8\columnwidth}{!}{%
\begin{tabular}{ll}
\toprule
\textbf{Variable} & \textbf{Search Range} \\
\midrule
$\text{LR}_{\text{stage1}}$ & $\{3\times 10^{-4}, 1\times 10^{-3}\}$ \\
$\text{LR}_{\text{stage2}}$ & $\{1\times 10^{-4}, 3\times 10^{-4}, 1\times 10^{-3}\}$ \\
$\text{WD}_{\text{stage1}}$ & $\{0, 0.01, 0.05\}$ \\
$\text{WD}_{\text{stage2}}$ & $\{0, 1\times 10^{-4}, 1\times 10^{-3}\}$ \\
$\lambda_1$ (Path) & $[0.0, 0.4]$ (step $0.05$) \\
$\lambda_2$ (Rebound) & $[0.0, 0.4]$ (step $0.05$) \\
$\lambda_{\text{warmup}}$ & $\{0, 0.3\}$ \\
\bottomrule
\end{tabular}%
}
\end{table}

\section{Additional Experimental Results}\label{sec:appendix_results}
\subsection{Case of Hallucination Dynamics}
\label{sec:appendix_case_studies}

\begin{table*}[htbp]
\centering
\small
\renewcommand{\arraystretch}{1.2}
\begin{tabularx}{\textwidth}{l l X}
\toprule
\textbf{Dataset} & \textbf{Type} & \textbf{Question \& Response} \\
\midrule
\textit{HotpotQA} & \textcolor{blue}{Factual} & \textbf{Q:} Are Manhattan West and Singer Building both projects in New York? \\
& & \textbf{R:} \textcolor{blue}{Yes}. \\
\cmidrule{2-3}
& \textcolor{red}{Hallucination} & \textbf{Q:} What State has a Disney Resort \& Spa that is a beachside hotel, resort and vacation destination offering complimentary children's activities and programs and that Djuan Rivers was a General Manager at? \\
& & \textbf{R:} The Disney Resort and Spa where Djuan Rivers served as General Manager is located in \textcolor{red}{Florida}. \\
\midrule
\textit{TriviaQA} & \textcolor{blue}{Factual} & \textbf{Q:} Where did the Shining Path terrorists operate? \\
& & \textbf{R:} The Shining Path terrorists operated in \textcolor{blue}{Peru}. \\
\cmidrule{2-3}
& \textcolor{red}{Hallucination} & \textbf{Q:} Who had a 70s No 1 hit with Kiss You All Over? \\
& & \textbf{R:} \textcolor{red}{The Village People} had a 70s No 1 hit with ``Kiss You All Over.'' \\
\bottomrule
\end{tabularx}
\caption{Examples of factual and hallucinated samples. Each dataset displays a Factual sample followed by a Hallucination. Factual samples exhibit consistent convergence, whereas hallucinations trigger detectable instability in denoising dynamics.}
\label{tab:case_study}
\end{table*}

Table \ref{tab:case_study} details the four samples discussed in our Introduction. For factual cases like the New York projects and the Shining Path, the denoising entropy steadily decreases as the model converges on the correct facts in the late-stage. In contrast, hallucinations show clear instability. In the Disney case, the model is biased toward ``Florida'' maybe due to the keyword ``Disney,'' but the specific constraint ``Djuan Rivers'' causes an entropy rebound when the model fails to resolve the conflict. Similarly, the hallucination for the 70s hit shows a late-stage bounce, indicating a breakdown in factual retrieval. These examples confirm that the trajectory shape is a more reliable signal for detection than absolute entropy values.

\subsection{Distinguishability of Statistical Evidence Trajectories}

\begin{figure}[ht]
    \centering
    \includegraphics[width=0.95\columnwidth]{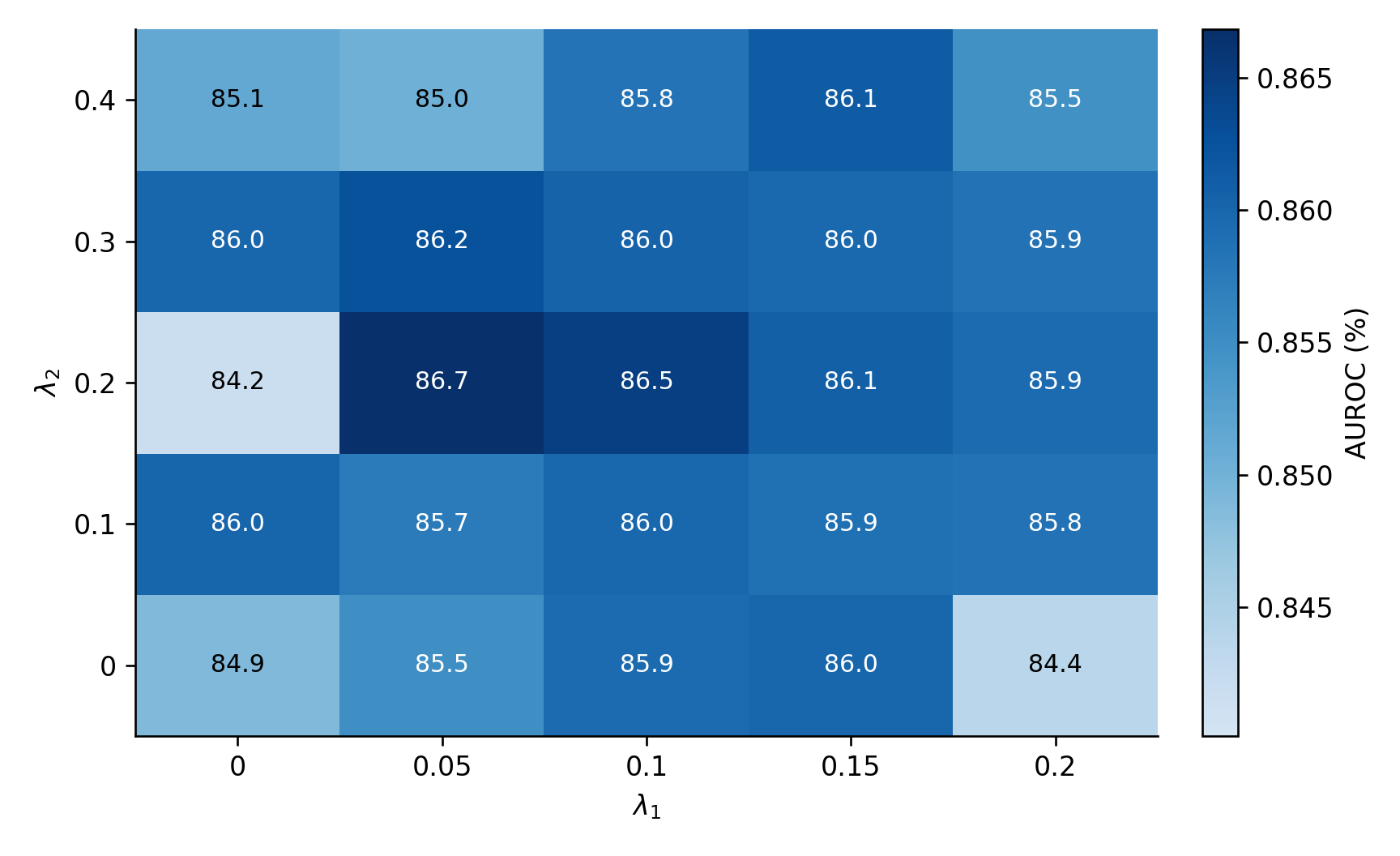}
    \caption{AUROC heatmap for varying $\lambda_1$ (Path) and $\lambda_2$ (Rebound) on TriviaQA. The peak region validates the synergistic effect of both dynamical regularizers.}
    \label{fig:lambda_heatmap}
\end{figure}

\begin{figure}[ht]
    \centering
    \includegraphics[width=0.85\columnwidth]{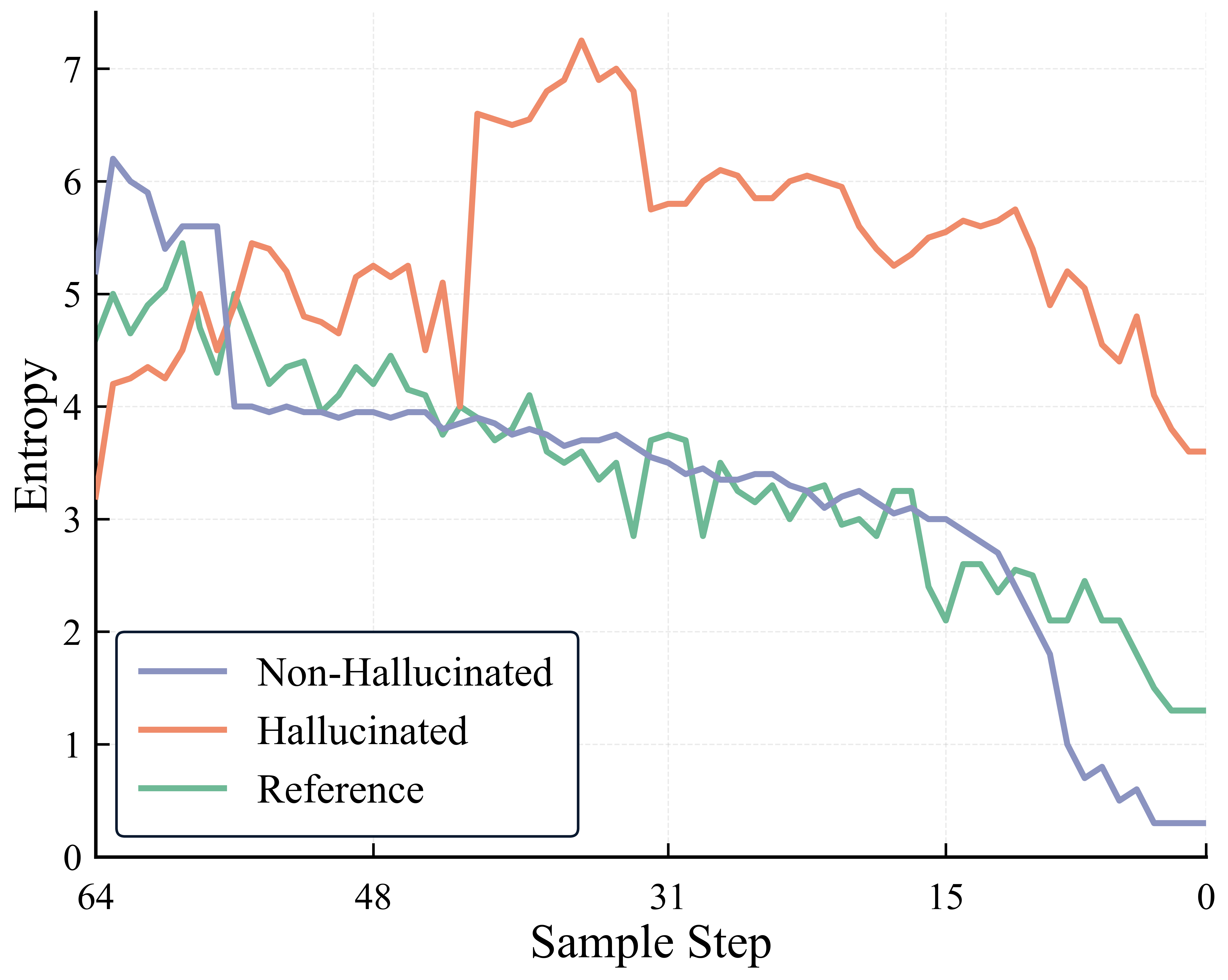}
    \caption{Empirical denoising dynamics of $a_t$ (observed) and $\hat{a}_t$ (reference) at 64 steps on HotpotQA. }
    \label{fig:dynamics_64}
\end{figure}

We test how our metrics separate factual answers from hallucinations in Figures \ref{fig:global_dist}--\ref{fig:trivia_dist}. 
As shown in Figure \ref{fig:global_dist}, our filtering method clarifies the uncertainty data. This creates a clear difference between facts and hallucinations, whereas raw data often overlaps. 
Specific tests in Figures \ref{fig:hotpot_dist} and \ref{fig:trivia_dist} show a common pattern: factual paths smoothly go down, while hallucinations bounce at the end. 

\subsection{Sensitivity of $\lambda_1$ and $\lambda_2$}
\label{sec:appendix_lambda_sensitivity}

Figure \ref{fig:lambda_heatmap} shows how $\lambda_1$ (Path) and $\lambda_2$ (Rebound) work together on TriviaQA. 
The heatmap indicates that the model works best when both $\lambda_1$ and $\lambda_2$ are used (around $0.1$ to $0.3$). This confirms that tracking the whole path and checking the late-stage bounce are both important. 
The model performs well across a wide range of values, showing it is not overly sensitive to specific parameter settings.

\subsection{Denoising Dynamics at 64 Steps}
\label{sec:appendix_dynamics_64}

Figure \ref{fig:dynamics_64} compares the real paths $a_t$ and the reference curves $\hat{a}_t$ at 64 steps. 
Factual answers follow the reference curve closely, but hallucinations show a clear bounce. 
These results prove that the shape of the curve is the most important indicator of an error, regardless of the exact values.

\end{document}